\documentclass{article}

\usepackage[eandd, final]{neurips_2026}

\usepackage[utf8]{inputenc} 
\usepackage[T1]{fontenc}    
\usepackage{hyperref}       
\usepackage{url}            
\usepackage{booktabs}       
\usepackage{amsfonts}       
\usepackage{nicefrac}       
\usepackage{microtype}      
\usepackage{xcolor}         

\usepackage[normalem]{ulem}
\usepackage[acronym]{glossaries}
\usepackage{xspace}

\usepackage{graphicx}
\usepackage{pythonhighlight}
\usepackage{bold-extra}
\usepackage{enumitem}
\usepackage{listings}
\usepackage{color}
\usepackage{lastpage} 
\usepackage{cleveref}
\usepackage{comment}
\usepackage{multirow}
\usepackage{tabularx}
\usepackage{wrapfig}
\usepackage{makecell}
\usepackage{dsfont}
\usepackage{bm}        

\pdfmapline{+poppins < Poppins-SemiBold.ttf <T1-WGL4.enc}
\DeclareFontFamily{T1}{poppins}{}
\DeclareFontShape{T1}{poppins}{m}{n}{<->s*[1.0]poppins}{}

\newcommand\poppinsfont[1]{{\usefont{T1}{poppins}{m}{n}#1}}
\newcommand{\titlequanda}{\poppinsfont{quanda}\xspace}
\newcommand{\quanda}{{\fontsize{8.0pt}{11pt}\poppinsfont{quanda}}\xspace}
\newcommand{\appquanda}{{\fontsize{7.5pt}{11pt}\poppinsfont{quanda}}\xspace}

\definecolor{codebg}{HTML}{FAFAF2}
\definecolor{codeframe}{HTML}{A0A598}
\definecolor{codekeyword}{HTML}{4F7E78}  
\definecolor{codestring}{HTML}{E41517}   
\definecolor{codecomment}{HTML}{6A737D}  
\definecolor{codenumber}{HTML}{A0A0A0}   
\definecolor{codeemph}{HTML}{1B7A52}     

\lstdefinestyle{mypython}{
  language=Python,
  basicstyle=\ttfamily\footnotesize,
  keywordstyle=\color{codekeyword}\bfseries,
  stringstyle=\color{codestring},
  commentstyle=\color{codecomment}\itshape,
  emphstyle=\color{codeemph},
  emph={self,True,False,None},
  numbers=none,
  numberstyle=\tiny\color{codenumber},
  numbersep=8pt,
  backgroundcolor=\color{codebg},
  frame=single,
  rulecolor=\color{codeframe},
  framerule=0.2pt,
  framesep=6pt,
  framexleftmargin=4pt,
  xleftmargin=14pt,
  xrightmargin=4pt,
  showstringspaces=false,
  breaklines=true,
  breakatwhitespace=true,
  keepspaces=true,
  columns=fullflexible,
  upquote=true,
  tabsize=4,
  aboveskip=0.8em,
  belowskip=0.8em,
}
\definecolor{dkgreen}{rgb}{0,0.6,0}
\definecolor{gray}{rgb}{0.5,0.5,0.5}
\definecolor{mauve}{rgb}{0.58,0,0.82}

\newcommand{\iconm}{\raisebox{-0.05em}{\includegraphics[height=0.8em]{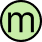}}}
\newcommand{\iconf}{\raisebox{-0.05em}{\includegraphics[height=0.8em]{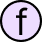}}}

\newacronym{ai}{AI}{artificial intelligence}
\newacronym{ml}{ML}{machine learning}
\newacronym{XAI}{XAI}{eXplainable artificial intelligence}

\title{\titlequanda: An Interpretability Toolkit for Training Data Attribution Evaluation}

\author{%
  Dilyara Bareeva$^{1}$ \quad Galip \"{U}mit Yolcu$^{1}$ \quad Anna Hedstr\"{o}m$^{2,3,4,5}$ \quad Niklas Schmolenski$^1$ \\
  \textbf{Thomas Wiegand}$^{1,5,6}$ \quad \textbf{Wojciech Samek}$^{1,5,6\dagger}$ \quad \textbf{Sebastian Lapuschkin}$^{1,7\dagger}$ \\
  $^1$Fraunhofer HHI \quad $^2$ETH AI Center  \quad
  $^3$Solaris Research  \quad  $^4$ATB Potsdam  \\
  $^5$TU Berlin  \quad 
  $^6$BIFOLD  \quad 
  $^7$TU Dublin \\
\textsuperscript{$\dagger$} corresponding authors:  \\\texttt{\{wojciech.samek,sebastian.lapuschkin\}@hhi.fraunhofer.de}\\
}

\begin{document}

\maketitle

\begin{abstract}
Training data attribution (TDA) methods estimate the influence of individual training samples on a model's predictions, offering a principled lens for interpreting neural networks. Despite rapid methodological progress, TDA evaluation remains scattered across papers and rarely reproducible, making fair comparison between methods challenging. We introduce \quanda, a \texttt{Python} toolkit that standardizes TDA evaluation. \quanda features a comprehensive collection of evaluation metrics and ready-to-use benchmarks spanning image classification, text classification, and language modeling, alongside a uniform interface for integrating diverse TDA implementations. We showcase \quanda through a comparative study of existing TDA methods and find that no approach excels uniformly across metrics, with clear trade-offs between ground-truth faithfulness and downstream task performance. By consolidating the tooling around TDA, \quanda equips the community to evaluate attribution methods rigorously and reproducibly. The toolkit is thoroughly tested, documented, and available as an open-source library on PyPI and at \url{https://github.com/dilyabareeva/quanda}.

\end{abstract}

\section{Introduction}
As neural networks become increasingly complex and widely adopted, the field of \emph{Explainable AI} (XAI) has emerged with the goal of making the rationales behind the predictions of these black-box models understandable to humans~\citep{longo2024explainable}. Early research in XAI predominantly focused on feature attribution methods, which highlight features in the input space that are responsible for a specific prediction~\citep{simonyan2014deepinsideconvolutionalnetworks, bach_pixel-wise_2015, scott2017unified, sundararajan2017axiomatic}. However, limitations in these methods' reliability~\citep{adebayo2018sanity, Gghorbani2019interpretation, bilodeau2024impossibility} and their lack of informativeness~\citep{rudin2019stop,achtibat2023from} have prompted a surge in alternative approaches. These include \emph{mechanistic interpretability}~\citep{bereska2024mechanisticinterpretabilityaisafety} methods, which aim to reverse-engineer neural networks by identifying and understanding the features and their connections within the model. In parallel, \emph{concept-based} explainability approaches seek to explain model predictions via high-level, human-understandable concepts that the model uses during inference~\citep{poeta2023conceptbasedexplainableartificialintelligence}.

Our work puts its focus on the interpretability approach known as \emph{training data attribution} (TDA)~\citep{hammoudeh_training_2024}. Broadly, TDA methods aim to relate a model's inference on a specific sample to its training data~\citep{koh2017understanding,yeh2018representer,pruthi2020estimating,park2023trak,bae2024training}. We denote by $\mathcal{D} = (\mathsf{z}_1, \dots, \mathsf{z}_n) \in \mathcal{Z}^n$ an ordered set composed of $n$ training samples, where each sample $\mathsf{z}_i=\left(\bm{x}_i,y_i\right)\in \mathcal{Z}$. Here, $\bm{x}_i$ represents the model input and $y_i$ represents the target variable. The space $\mathcal{Z}$ corresponds to the set of all possible input-target pairs, and the targets' domain can vary depending on a task. For instance, in a classification setting, $y_i$ typically represents a label. Then, given a test sample $\mathsf{z} = \left(\bm{x}, y\right)$, a data attribution method defines a function $\tau: \mathcal{Z} \times \mathcal{Z}^{n} \rightarrow \mathbb{R}^n$ that assigns a real-valued \emph{attribution score} to each training sample $\mathsf{z}_i$. The attributed variable is typically the model output $f\left(\mathsf{z}; \theta\right)$ or the model training loss $\mathcal{L}\left(\mathsf{z}; \theta\right)$ on sample $\mathsf{z}$. The attribution score, depending on the method, can be referred to as the influence~\citep{koh2017understanding, pruthi2020estimating}, importance~\citep{park2023trak, bae2024training}, representer value~\citep{yeh2018representer}, or similarity~\citep{caruana_case-based_1999, hanawa2021evaluation}.

Some TDA methods aim to estimate the \emph{counterfactual effects} of modifying the training dataset on model output or loss~\citep[]{koh2017understanding, koh2019on, park2023trak}. A classic approach for such estimation is leave-one-out (LOO) retraining~\citep{cook1982residuals}. LOO quantifies the influence of a sample by retraining the model after removing it from the training dataset and measuring the resulting change in the model's loss. Other methods leverage interpretable surrogates~\citep{yeh2018representer, yolcu2025sparse} or kernel methods~\citep{khanna2019interpreting}, identify training samples that are deemed similar to the test sample by the model~\citep{caruana_case-based_1999,hanawa2021evaluation}, or track the influence of training data points throughout the training process~\citep{pruthi2020estimating}.

Beyond facilitating model interpretability~\citep{pezeshkpour-etal-2021-empirical,naujoks2024pinnfluenceinfluencefunctionsphysicsinformed}, TDA has been proposed to be used in a variety of applications, such as model debugging~\citep{koh2017understanding,yeh2018representer,guo-etal-2021-fastif}, data pruning~\citep{marion2023when,yang2023dataset}, data selection~\citep{chhabra2024what, engstrom2024dsdm}, subsampling~\citep{ting2018optimal}, machine unlearning~\citep{WarPirWreRie20}, and fact tracing~\citep{akyurek-etal-2022-towards}.

\begin{figure} 
  \centering
  \includegraphics[width=0.9811320755 \textwidth]{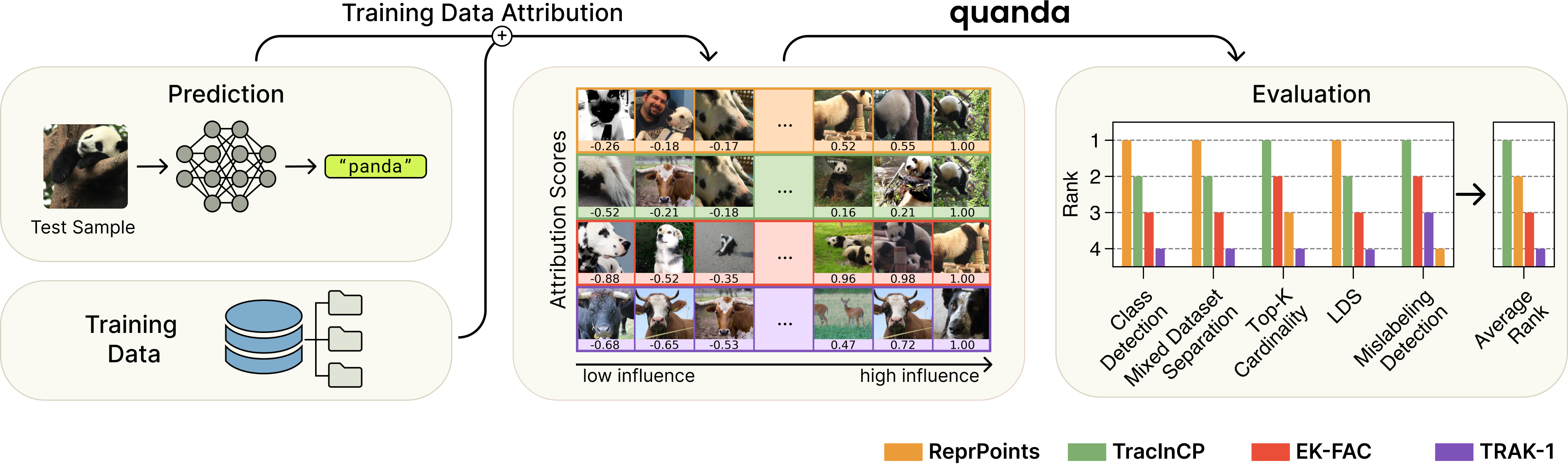}
  \caption{Overview of exemplary benchmarking results facilitated by \quanda, instantiated on the AwA2/ResNet-50 setting. The figure depicts the training data attribution (TDA) evaluation pipeline: each TDA method generates attribution scores for test samples, and \quanda computes metric scores, assessing various aspects of attribution quality across different evaluation strategies. Each metric provides insight into a specific dimension of the attributor's performance, resulting in a comprehensive analysis of the strengths and weaknesses of each method. The full quantitative results for this setting are reported in~\Cref{fig:eval-image-classification}, and the experimental details are given in~\Cref{exp_details}.
  }
  \label{fig:overview}
\end{figure}

Although there are various demonstrations of TDA's potential for interpretability and practical applications, a comprehensive and standardized evaluation framework remains lacking.
The gap hinders a detailed understanding of the strengths and limitations of different TDA techniques across various contexts. Furthermore, the limited adoption and limited popularity of TDA methods~\citep{nguyen_exploring_2023} may in part be attributed to the absence of user-friendly, efficient, and ready-to-use tools for applying and evaluating different TDA techniques. To enhance the reliability of TDA-based explanations and task outcomes, researchers urgently need standardized implementations of various TDA explainers and unified, comprehensive evaluation tools. We address these research gaps with the release of the open-source \texttt{Python} package \quanda, implemented for \texttt{PyTorch}~\citep{paszke_pytorch_2019}, offering the following contributions:

\begin{itemize}
  \item A \emph{standardized interface} for multiple TDA methods, which are currently scattered across different repositories, simplifying the application and comparison of various methods.
  \item Implementations of several \emph{evaluation metrics} from the literature, enabling the assessment of explanation quality and downstream-task performance.
  \item Unified \emph{benchmarking tools} and standardized \emph{benchmarking assets}, facilitating metrics-based evaluation of TDA methods in controlled environments.
\end{itemize}
\section{Related Work} \label{related_work}

The need to assess the reliability of explanations and select the most suitable TDA method for a given intent and application raises the critical question of how TDA methods should be effectively evaluated. As some of the methods are designed to approximate LOO effects, \emph{ground truth} can often be computed for TDA evaluation~\citep[e.g.,][]{koh2017understanding,koh2019on, basu2021influence, park2023trak}. However, this counterfactual ground truth approach requires retraining the model multiple times on different subsets of the training data, which quickly becomes computationally expensive. Furthermore, this ground truth is shown to be dominated by noise in practical deep learning settings, due to the inherent stochasticity of a typical training process~\citep{nguyen_exploring_2023}. The model parameters after optimization are often sensitive to hyperparameters and initialization, introducing further uncertainty, as discussed in Appendix A of~\citet{bae2024training}.

Alternative approaches make evaluation feasible at scale. \emph{Downstream task evaluators} assess the utility of a TDA method within the context of an end-task, such as model debugging or data selection~\citep[e.g.,][]{koh2017understanding, yeh2018representer, pruthi2020estimating}. \emph{Heuristics} evaluate whether a TDA explanation meets certain expected properties, such as its dependence on model weights or the diversity of top-attributed samples across different test samples~\citep[e.g.,][]{barshan2020relatif,hanawa2021evaluation,singla2023simpleefficientbaselinedata}. However, these evaluation strategies, developed independently by various researchers, often rely on distinct assumptions and implementational nuances that can affect their outcomes. As a result, the evaluation results reported in independent studies are frequently not directly comparable.

Over the past years, a growing ecosystem of software libraries for TDA methods has emerged. One of the earliest efforts, \texttt{pyDVL}~\citep{pyDVL} provides influence
function implementations with a focus on data valuation~\citep{sim2022data}. \texttt{Influenci\ae}~\citep{picard2024influenciae} implements influence-based methods in \texttt{TensorFlow}~\citep{abadi_tensorflow_2016}. \texttt{dattri}~\citep{dattri} has introduced a broader, TDA-oriented \texttt{Python} framework covering a wide range of attribution methods; however, less than a handful of evaluation metrics are included. \texttt{DATE-LM}~\citep{datelm} provides a reusable benchmarking suite for large language models (LLMs), though it is not packaged as a general-purpose \texttt{Python} library. None of these libraries makes it straightforward to extend evaluation to TDA methods they do not natively support, and to this day there remains no unified, open-source framework for standardized evaluation of data attribution across tasks, modalities, and implementations.

\quanda addresses the need for an open-source, standardized evaluation library for data attribution across different tasks. In~\Cref{tab:library_comparison}, we provide a short summary of the current TDA-evaluation landscape, highlighting the gap that \quanda fills. 
In this respect, \quanda draws significant inspiration from previous work on feature attribution evaluation, notably \texttt{Quantus}~\citep{hedstroem2023quantus} and \texttt{OpenXAI}~\citep{agarwal2022openxai}. In the vein of these two libraries, \quanda was designed to unify disparate evaluation strategies and to provide benchmarking tools for systematic comparison of techniques.

\begin{table}[h!]
\centering
\caption{Comparison of existing data attribution libraries with prominent evaluation features. \textit{Modalities} indicates the tasks supported among image classification, text classification, and language modeling (\texttt{dattri} additionally features music generation). \textit{Metric Categories} reports coverage across the three metric families considered in this work: heuristics, ground-truth metrics, and downstream-task evaluators. \textit{Explainer Interface} indicates whether the library exposes a public interface for user-defined attribution methods.}
\label{tab:library_comparison}
\begin{tabular}{lcccc}
\toprule
Library & \shortstack{\#\\Metrics} & \shortstack{\#\\Metric Categories} & \shortstack{\#\\Modalities} & \shortstack{Explainer\\Interface} \\
\midrule
\quanda{} (ours)                         & $11$ & 3 & 3 & $\checkmark$ \\
\texttt{dattri}~\citep{dattri}           &  $4$ & 2 & 3 & $\times$ \\
\texttt{DATE-LM}~\citep{datelm}          &  $5$ & 1 & 1 & $\times$ \\
\bottomrule
\end{tabular}

\end{table}

\section{Library Overview} \label{library}

\quanda is designed to facilitate the evaluation of TDA methods for practitioners, model developers, and researchers. ~\Cref{fig:overview} illustrates the typical workflow: given a trained model and its training data, attribution methods produce per-example scores, which \quanda transforms into a set of evaluation metric scores that expose strengths, weaknesses, and trade-offs across methods.

The following sections describe the library's core components, key features, and API design. \quanda currently focuses on \emph{post-hoc} data attribution and supports attributing decisions of a single model, as opposed to computing average attributions for different instantiations of the model architecture as sometimes done in the literature~\citep{k2021revisitingmethodsfindinginfluential,park2023trak}.

\subsection{Library Design}
\begin{figure} 
  \centering
  \includegraphics[width=0.9811320755 \textwidth]{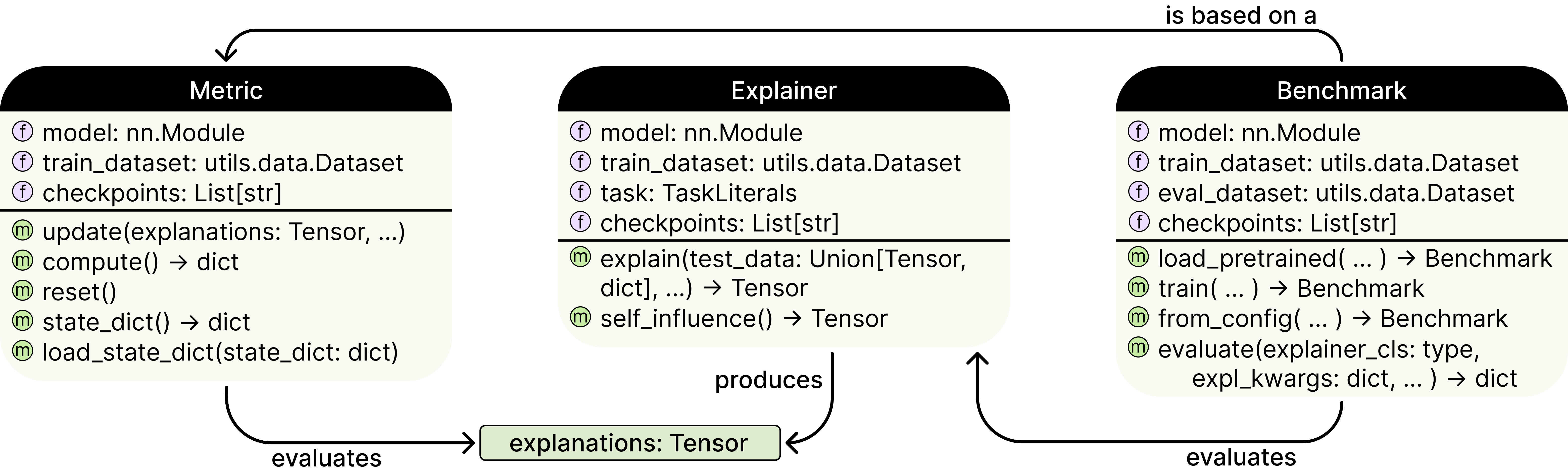}
  \caption{Illustration of main \quanda components. Each block represents a base class. The class fields are listed in the top section, indicated with the icon \iconf, while the class methods are listed in the bottom, indicated with \iconm. The leading \texttt{torch} is omitted for \texttt{PyTorch} types. The arrows explain the functions of and relations between individual components. Specifically, while \texttt{Explainer} and \texttt{Metric} classes relate indirectly through the generated explanations, the \texttt{Benchmark} class internally utilizes a \texttt{Metric} instance to evaluate the \texttt{Explainer} class, handling the generation of explanations.}
  \label{fig:components}
\end{figure}

The design philosophy of \quanda emphasizes creating modular interfaces that represent distinct functional units. The three main components of \quanda are explainers, evaluation metrics, and benchmarks. Each component is implemented as a base class that defines the minimal functionalities required to implement a new instance. The design of the base classes prioritizes easy extension and integration with other components. This allows users, for instance, to evaluate a novel TDA method by simply wrapping their implementation to conform to the base explainer interface. \Cref{fig:components} illustrates the main functionality of different components and lists the core class fields and methods.

\begin{description}[style=unboxed,leftmargin=0cm]

\item[Explainers] An \texttt{Explainer} is a class representing a single TDA method. An \texttt{Explainer} instance maintains information about the specific model architecture, model weights, training dataset, and in some cases the training hyperparameters, such as the loss function.
Many TDA methods require a computationally expensive initialization step. This step is performed once, when an \texttt{Explainer} is instantiated, and the attribution of a test batch is then performed on-demand with the \texttt{explain} method. In addition, the \texttt{self\_influence} method returns a vector of self-influences, i.e., attributions of each training sample to itself~\citep{koh2017understanding}.

\item[Metrics] A metric summarizes the performance and reliability of a TDA method in a compact representation, typically as a single number. Three categories of \texttt{Metric} classes can be found in \quanda: \texttt{ground\_truth}, \texttt{downstream\_eval} and \texttt{heuristics}. To provide flexibility and efficiency, \quanda adopts a \textit{stateful} \texttt{Metric} design that supports the incremental addition of new test batches via the \texttt{update} method, where applicable. This allows the user to directly evaluate precomputed explanations, or to conduct the evaluation process alongside the explanation process. The final metric score is obtained by calling the \texttt{compute} method after the \texttt{Metric} instance has been updated with the explanations. 

\item[Benchmarks] A \quanda \texttt{Benchmark} bundles a model architecture and weights, training and evaluation datasets, and a metric with its parameters, similar to the definition of \citet{raji2021ai}. We provide ready-made suites spanning a range of metrics, architectures, and tasks, with all assets pre-computed and configured, hosted on \texttt{Hugging Face}~\citep{hf_datasets,hf_transformers} and accessible via \texttt{load\_pretrained} given a benchmark ID. \texttt{Benchmark} instances can also be initialized from a YAML configuration via \texttt{from\_config}, which assumes weights and dataset metadata are already available either on \texttt{Hugging Face} or in a local cache directory. To construct a benchmark from scratch from a YAML configuration, users can invoke \texttt{train} or \texttt{train\_and\_push\_to\_hub} to additionally upload the ready-made benchmark to \texttt{Hugging Face}. Once initialized, an \texttt{Explainer} with a given hyperparameter configuration can be quantitatively assessed by calling \texttt{evaluate}, which computes the associated metric score. The \texttt{explain} method is featured to generate and cache attributions, hashed by their evaluation parameters, for reuse across benchmarks that share the evaluation setting (i.e., model and evaluation dataset).  The following code snippet contains an example of how \quanda can be used to evaluate a TDA method with a ready-made benchmark. 
\end{description}

\vspace{-0.5em}

\begin{lstlisting}[style=mypython,label={lst:quanda},caption={Evaluating an explainer with \quanda's ready-made benchmark.}]
import quanda

class_detect = quanda.ClassDetection.load_pretrained(bench_id="mnist_class_detection", cache_dir=cache_dir)
    
trak_expl_kwargs = {"model_id": "test_model", "cache_dir": cache_dir, "proj_dim": 512, "seed": 42}

result = class_detect.evaluate(explainer_cls=quanda.TRAK, expl_kwargs=trak_expl_kwargs, batch_size=batch_size, max_eval_n=16)["score"]

print("Class Detection Score (TRAK):", result)
\end{lstlisting}
\vspace{-0.5em}

\subsection{Maintaining Library Quality and Accessibility}

We ensure high code quality through thorough integration and unit testing with \texttt{pytest}, enforcing $95\%$ test coverage. Type safety is enforced with the static type checker \texttt{mypy}, and style consistency with PEP 8 conventions is maintained through the linter and formatter \texttt{ruff}. Each pull request triggers a continuous integration (CI) pipeline on GitHub that runs the full test suite and checks for coverage, type correctness, linting, docstring completeness, and validity of documentation code.

The \quanda toolkit leverages widely-used libraries such as \texttt{PyTorch Lightning}, \texttt{Hugging Face Datasets} and \texttt{Transformers} to ensure seamless integration into users' pipelines while avoiding the reimplementation of functionality that is already available in established libraries. The \texttt{Explainer} classes provide wrappers for existing \texttt{PyTorch} implementations of TDA methods, including those from \texttt{Captum}~\citep{kokhlikyan_captum_2020}, the official \texttt{TRAK} implementation~\citep{park2023trak}, \texttt{Kronfluence}~\citep{grosse2023studyinglargelanguagemodel}, \texttt{dattri}~\citep{dattri}, and the official Representer Points repository~\citep{yeh2018representer}. We plan to expand the number of supported TDA wrappers.

In designing the \quanda library, the primary focus is to ensure ease of use by providing clear and consistent APIs. Comprehensive documentation, installation guides, and extensive tutorials are provided to help users navigate the library for attribution, evaluation, and benchmarking. The library documentation is available at the following URL: \url{https://quanda.readthedocs.io}.

The \quanda source code is released as a GitHub repository and a PyPI package under the open-source MIT license. The benchmark assets are hosted on \texttt{Hugging Face} under MIT license and can be loaded directly through \quanda. We actively welcome bug reports and contributions through GitHub, supported by detailed contributor guidelines to lower the barrier to community involvement.
\section{Evaluation Metrics} \label{metrics_desc}

The metrics in \quanda draw on both seminal TDA papers~\citep{koh2017understanding,pruthi2020estimating,park2023trak} and dedicated work on TDA evaluation and application~\citep{hanawa2021evaluation,hammoudeh2022identifying,chang2025scalable}. We briefly describe the metrics featured in \quanda below, grouping them into ground-truth metrics, downstream-task metrics, and heuristics. 
Details on the notations are given in~\Cref{notation}.

\subsection{Ground Truth Metrics}
Many TDA methods are grounded in approximating the counterfactual effect of omitting training points on model outputs or loss, suggesting the \emph{ground-truth-based} evaluation strategy: directly measuring how well a method's attributions predict the effect of retraining on training subsets.

\paragraph{Linear Datamodeling}
We currently implement the most prominent representative of the \emph{ground-truth} family, the Linear Datamodeling Score (LDS)~\citep{ilyas2022datamodels,park2023trak}, which has become the de facto standard against which new TDA methods are benchmarked~\citep{bae2024training,ilyas2025magicnearoptimaldataattribution,sun2026enhancing}. Let $f(\mathsf{z};\theta^*(\mathcal{D}'))$ denote the output of a model trained on a subset $\mathcal{D}'\subseteq\mathcal{D}$, evaluated at $\mathsf{z}$. Assuming attributions are \emph{additive}, the attribution-based prediction of the output of a subset model is
\begin{equation}
    g_\tau(\mathsf{z}, \mathcal{D}'; \mathcal{D}) = \sum_{i\,:\,\mathsf{z}_i\in\mathcal{D}'} \tau(\mathsf{z}, \mathcal{D})_i.
\end{equation}
For $m$ random subsets $\{\mathcal{D}'_j\}_{j=1}^{m}$ with $\mathcal{D}'_j\subseteq\mathcal{D}$, LDS averages the Spearman rank correlation $\rho$ between predicted and actual outputs over the test set:
\begin{equation}
    \mathrm{LDS}(\tau) = \frac{1}{|\mathcal{D}_{\text{test}}|}\sum_{\mathsf{z}\in\mathcal{D}_{\text{test}}}\rho\!\left(\{g_\tau(\mathsf{z},\mathcal{D}'_j;\mathcal{D})\}_{j=1}^{m},\;\{f(\mathsf{z};\theta^*(\mathcal{D}'_j))\}_{j=1}^{m}\right).
\end{equation}

\subsection{Downstream-Task Evaluators}
As discussed in~\Cref{related_work}, \emph{downstream-task evaluators} offer a practical alternative to \emph{ground-truth} metrics, as they are often significantly cheaper to compute and directly tied to real-world applications.

\paragraph{Class and Subclass Detection}
A natural intuition is that the top proponents of a test sample, i.e., highest-scoring training samples, should be similar to it. Following \citet{hanawa2021evaluation}, we test this through two related metrics: Class Detection, which asks whether the top-attributed training points share the predicted class of the test sample, and Subclass Detection, which, for a model trained on randomly grouped superclasses, tests whether attributions recover the original subclass of the test sample. Both report
\begin{equation}
    \mathrm{ClassDet}_s(\tau)=\frac{1}{|\mathcal{D}_{\text{test}}|}\sum_{\mathsf{z}\in\mathcal{D}_{\text{test}}}\frac{1}{s}\sum_{i=1}^{s}\mathds{1}\!\left[c(\mathsf{z})=y_{\pi^\tau_\mathsf{z}(i)}\right],
\end{equation}
where $\mathcal{D}_{\text{test}}$ is the evaluation set, $\pi^\tau_\mathsf{z}$ is the permutation that sorts training points in descending order of attribution scores derived from $\tau$ for sample $\mathsf{z}$, $c(\cdot)\in[C]$ is the comparison target (typically the predicted class for Class Detection and the ground-truth subclass label for Subclass Detection), $y_j$ is the label of the $j$-th training sample (its original subclass label for Subclass Detection), and $s$ is the number of considered proponents. Higher metric values indicate better (sub)class recovery. We classify these as \emph{downstream-task evaluators} in \quanda, as they have been suggested to measure the practical utility of TDA methods in identifying same-class training data \citep{kwon2024datainf,li-etal-2025-influence}. Importantly, similarity-based criteria are not aligned with the counterfactual notion of attribution quality, as divergent-class training samples may still exert a large LOO effect through training-process idiosyncrasies.

\paragraph{Mislabeling Detection}\label{mislabeling_detection}
The use of TDA to identify and fix mislabeled training examples was proposed by \citet{koh2017understanding} and has since become a standard evaluation criterion~\citep{yeh2018representer,pruthi2020estimating,khanna2019interpreting}. The intuition is that mislabeled samples cannot be classified by relying on similar same-class examples and must instead be memorized, leaving them with disproportionately strong influence on their own predictions. 
Given a model trained on data with labels of a random subset $\mathcal{M}\subset[n]$ corrupted, we rank training points by descending self-influence $\tau(\mathsf{z}_i, \mathcal{D})_i$ to obtain a permutation $\pi^\tau_{\text{self}}$. We then plot cumulative recall against rank position and report the normalized area under this curve (AUC), where higher values indicate that mislabeled points concentrate near the top of the ranking:
\begin{equation}
    \mathrm{MislabelDet}(\tau) = \mathrm{AUC}(\mathrm{Recall}), \qquad \mathrm{Recall}(r) = \frac{1}{|\mathcal{M}|}\sum_{i=1}^{r}\mathds{1}[\pi^\tau_{\text{self}}(i)\in\mathcal{M}].
\end{equation}

\paragraph{Shortcut Detection}
A \emph{shortcut} (or Clever-Hans effect) is a decision rule that fits the training distribution but fails under distribution shift~\citep{lapuschkin2019unmasking,geirhos2020shortcut}. Originally proposed for influence functions as \emph{domain-mismatch debugging}~\citep{koh2017understanding,yolcu2025sparse}, the test in \quanda applies a fixed perturbation $\omega(\cdot)$ (e.g., a small patch) to a subset $\mathcal{S}\subset[n]$ from a designated shortcut class $c'\in[C]$ and trains the model so that it learns to associate $\omega$ with $c'$. Evaluation is restricted to test points $\omega(\bm{x})$ whose ground-truth class differs from $c'$ that the retrained model classifies as $c'$, so the prediction is driven by $\omega$ and attributions computed for the predicted class should rank shortcut samples above clean ones. Following \citet{hammoudeh2022identifying}, we report the per-test-point AUPRC averaged over the eligible set $\mathcal{D}_{\text{test}}^\omega$, with higher values indicating better shortcut detection:
\begin{equation}
    \mathrm{ShortcutDet}(\tau)=\frac{1}{|\mathcal{D}_{\text{test}}^\omega|}\sum_{\mathsf{z}\in\mathcal{D}_{\text{test}}^\omega}\mathrm{AUPRC}\!\left(\{\tau(\mathsf{z},\mathcal{D})_i\}_{i=1}^n,\;\{\mathds{1}[i\in\mathcal{S}]\}_{i=1}^n\right).
\end{equation}

\paragraph{Fact Tracing}
Fact tracing~\citep{akyurek-etal-2022-towards} evaluates whether a TDA method can retrieve training examples that support factual predictions. By dataset construction, each query $\mathsf{z}\in\mathcal{D}_{\text{test}}$ comes with an entailment label $e^\mathsf{z}_i \in \{0, 1\}$ for each train sample, against which \quanda computes the metrics Mean Reciprocal Rank (MRR) and Recall@$k$:
\begin{equation}
    \mathrm{MRR}(\tau)=\frac{1}{|\mathcal{D}_{\text{test}}|}\sum_{\mathsf{z}\in\mathcal{D}_{\text{test}}}\frac{1}{\min\{i\,:\,e^\mathsf{z}_{\pi^\tau_\mathsf{z}(i)}=1\}},\quad
    \mathrm{Recall}@k(\tau)=\frac{1}{|\mathcal{D}_{\text{test}}|}\sum_{\mathsf{z}\in\mathcal{D}_{\text{test}}}\max_{i\le k}e^\mathsf{z}_{\pi^\tau_\mathsf{z}(i)}.
\end{equation}
For both metrics, higher values indicate better retrieval of entailing examples. Since fact-entailing samples do not necessarily exert the largest counterfactual influence and retraining-based estimation is very expensive at LLM scale, \citet{chang2025scalable} introduce the \emph{tail-patch} metric, which measures the effect on the target sequence probability of taking a single additional gradient step, averaged over the top-$k$ proponents. Letting $\theta^{+i}$ denote the parameters obtained from $\theta^*(\mathcal{D})$ by one step on $\mathsf{z}_i$ and $p(\mathsf{z};\theta)$ the model's probability of the target completion at $\mathsf{z}$,
\begin{equation}
    \mathrm{TailPatch}_k(\tau)=\frac{1}{|\mathcal{D}_{\text{test}}|}\sum_{\mathsf{z}\in\mathcal{D}_{\text{test}}}\frac{1}{k}\sum_{i=1}^{k}\bigl(p(\mathsf{z};\theta^{+\pi^\tau_\mathsf{z}(i)})-p(\mathsf{z};\theta^*(\mathcal{D}))\bigr).
\end{equation}

\subsection{Heuristics}
Rather than scoring against a ground truth or downstream task, the metrics in this group probe whether attributions satisfy properties any reasonable attributor should exhibit.

\paragraph{Mixed Datasets Separation}
\citet{hammoudeh2022identifying} first observe that many TDA methods fail in a simple setting where, after training on a mixture of two distinct data sources, the top proponents of test samples from one source often fail to include training samples from the same source. In our \quanda implementation, we include a small \emph{adversarial} component $\mathcal{D}_{\text{adv}}$ from a substantially different distribution into the base training data $\mathcal{D}_{\text{base}}$. All samples in $\mathcal{D}_{\text{adv}}$ are labeled to a single class $y_{\text{adv}}$ from $\mathcal{D}_{\text{base}}$, and the model is trained on $\mathcal{D}=\mathcal{D}_{\text{base}}\cup\mathcal{D}_{\text{adv}}$. Evaluation is performed on the set $\mathcal{D}_{\text{test}}^{\text{adv}}$ from the adversarial distribution classified as $y_{\text{adv}}$. The metric signals how well attribution scores separate the two data sources in the training set:
\begin{equation}
    \mathrm{MixedDatasets}(\tau)=\frac{1}{|\mathcal{D}_{\text{test}}^{\text{adv}}|}\sum_{\mathsf{z}\in\mathcal{D}_{\text{test}}^{\text{adv}}}\mathrm{AUPRC}\!\left(\{\tau(\mathsf{z},\mathcal{D})_i\}_{i=1}^n,\;\{\mathds{1}[\mathsf{z}_i\in\mathcal{D}_{\text{adv}}]\}_{i=1}^n\right).
\end{equation}

\paragraph{Model Randomization}
The model randomization check requires attributions to depend on the model parameters $\theta^*(\mathcal{D})$, not on inputs alone. Originally proposed by~\citet{adebayo2018sanity} in the context of feature attribution, it was adapted to TDA methods by~\citet{hanawa2021evaluation}. Let $\tilde\theta^{(1)},\dots,\tilde\theta^{(R)}$ denote $R$ independent reinitialized model weights and $\tau_\theta(\mathsf{z},\mathcal{D})$ the attribution produced with the model parametrized by $\theta$. The metric reports the average Spearman rank correlation between attributions under trained and randomized parameters, averaged over the $R$ randomizations and the test set. Scores close to zero indicate that the attributor depends more on the trained parameters:
\begin{equation}
    \mathrm{ModelRand}(\tau)=\frac{1}{R\,|\mathcal{D}_{\text{test}}|}\sum_{r=1}^{R}\sum_{\mathsf{z}\in\mathcal{D}_{\text{test}}}\rho\!\left(\tau_{\theta^*(\mathcal{D})}(\mathsf{z},\mathcal{D}),\;\tau_{\tilde\theta^{(r)}}(\mathsf{z},\mathcal{D})\right).
\end{equation}

\paragraph{Top-$k$ Cardinality}
A common failure mode of TDA methods is repeatedly assigning top scores to a very limited set of training samples, undermining their informativeness. Let $\mathrm{Top}_k^\tau(\mathsf{z})=\{\pi^\tau_\mathsf{z}(1),\dots,\pi^\tau_\mathsf{z}(k)\}$. Following~\citet{barshan2020relatif}, \quanda reports the size of the union of top-$k$ sets across the test set, normalized by its maximum possible size:
\begin{equation}
    \mathrm{TopKCard}_k(\tau)=\frac{\left|\bigcup_{\mathsf{z}\in\mathcal{D}_{\text{test}}}\mathrm{Top}_k^\tau(\mathsf{z})\right|}{k\cdot|\mathcal{D}_{\text{test}}|}.
\end{equation}
Very low scores indicate that top-$k$ proponents do not vary substantially across test points and signal a possible failure; otherwise, lack of variability is not necessarily unfaithful to the model.
\section{Benchmarking Training Data Attribution Methods} \label{benchmarking}

\begin{figure*}[t]
\centering
\includegraphics[width=\textwidth]{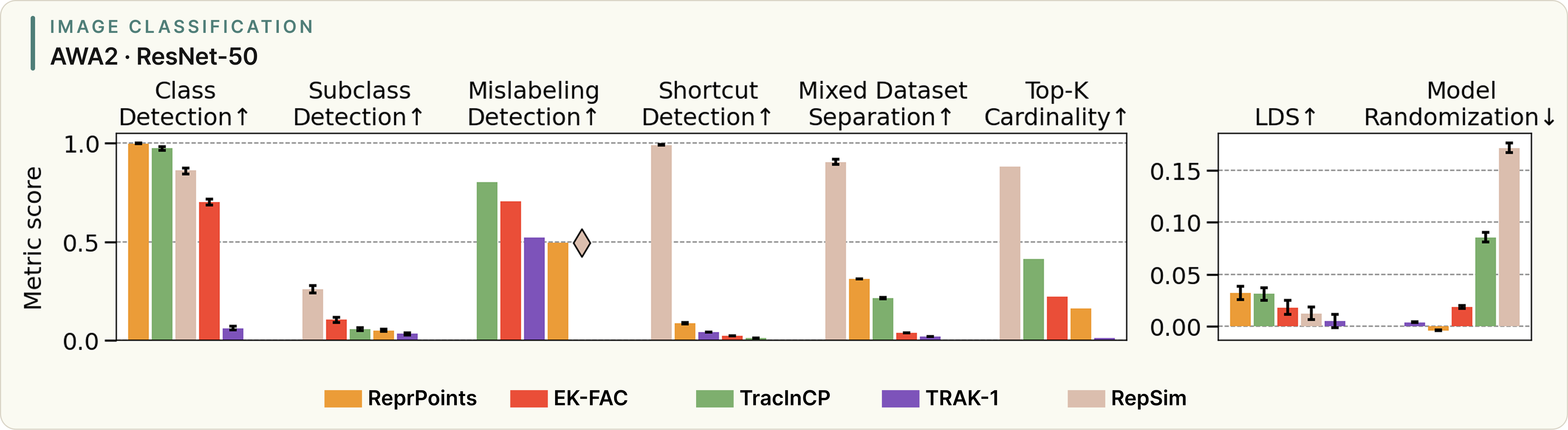}
\caption{Image-classification benchmarks (AwA2/ResNet-50). Bars give the score of each attribution method per benchmark (the best-performing configuration where a hyperparameter sweep was run; see~\Cref{exp-explainers}), sorted left-to-right by descending performance.  Arrows indicate whether a higher ($\uparrow$) or lower ($\downarrow$) value is preferable: the metric score in the left pane, the magnitude of the correlation-based metrics in the right pane. Error bars are $95\%$ bootstrap confidence intervals over per-test-sample scores ($1000$ resamples), shown only for benchmarks where per-sample resampling is meaningful. For reference, the random-attribution baseline (mean $\pm$ std over $41$ seeds) scores $0.025{\pm}0.002$ (Class), $0.025{\pm}0.002$ (Subclass), $0.495{\pm}0.002$ (Mislabeling), $0.014{\pm}0.000$ (Shortcut), $0.017{\pm}0.000$ (Mixed Datasets), $0.921{\pm}0.003$ (Top-$k$), $0.000{\pm}0.000$ (Model Randomization), and $0.000{\pm}0.004$ (LDS). RepSim's Mislabeling Detection score trivially collapses to $0.5$ and is drawn as a diamond marker.}
\label{fig:eval-image-classification}
\end{figure*}

To showcase the practical utility of \quanda, we conduct a broad empirical study spanning three modalities and three model families. The goal is to demonstrate that \quanda makes it straightforward to apply a heterogeneous set of metrics to a heterogeneous set of attributors on the same setting, and to reveal concrete observations about how current TDA methods behave across metrics. Each setting is available as a \quanda benchmark via \texttt{load\_pretrained}.
A study of this breadth would have been impractical without a unified interface and a library of ready-made benchmarks.

\subsection{Experimental Setup}
In the following, we report evaluation results of TDA methods across three representative settings: (i) image classification on Animals with Attributes 2 (AwA2)~\citep{xian2018zero} with ResNet-50~\citep{he_2016_cvpr}, following \citet{yolcu2025sparse}; (ii) text classification on the QNLI task from GLUE~\citep{wang-etal-2018-glue} with a fine-tuned BERT~\citep{devlin-etal-2019-bert}, following \citet{park2023trak}; and (iii) language modeling on T-REx~\citep{elsahar-etal-2018-rex} with GPT-2 small~\citep{radford2019gpt2}, following \citet{chang2025scalable}. Together, these settings span the principal tasks for which TDA is currently used, across both convolutional and transformer-based models.

In each setting, we evaluate a representative set of TDA methods appropriate to the task, architecture, and dataset scale. The applied metrics are determined by what is meaningful for the task: the AwA2 setting supports the full suite from~\Cref{metrics_desc} except fact tracing; the QNLI setting additionally omits Subclass and Shortcut Detection metrics, which are not meaningful for binary classification, and the GPT-2 setting uses the fact-tracing suite. As a sanity-check baseline, every setting also reports metric performance on a random attributor assigning i.i.d.\ Gaussian scores $\tau(\mathsf{z}, \mathcal{D})_i \sim \mathcal{N}(0, 1)$ independently of the model. Full specifications of the TDA methods, benchmark assets, and metrics, along with their hyperparameters and search grids, are listed in~\Cref{tda_appendix,benchmarks_appendix,exp_details}. Additional results on the other ready-to-use \quanda benchmarks, MNIST~\citep{lecun1998mnist} with LeNet-5~\citep{lecun_1998_lenet} and CIFAR-10~\citep{krizhevsky2009learning} with ResNet-9, are reported in~\Cref{app:extra_benchmark_results}.

\subsection{Evaluation Results}

\begin{figure*}[t]
\centering
\includegraphics[width=0.7330188679 \textwidth]{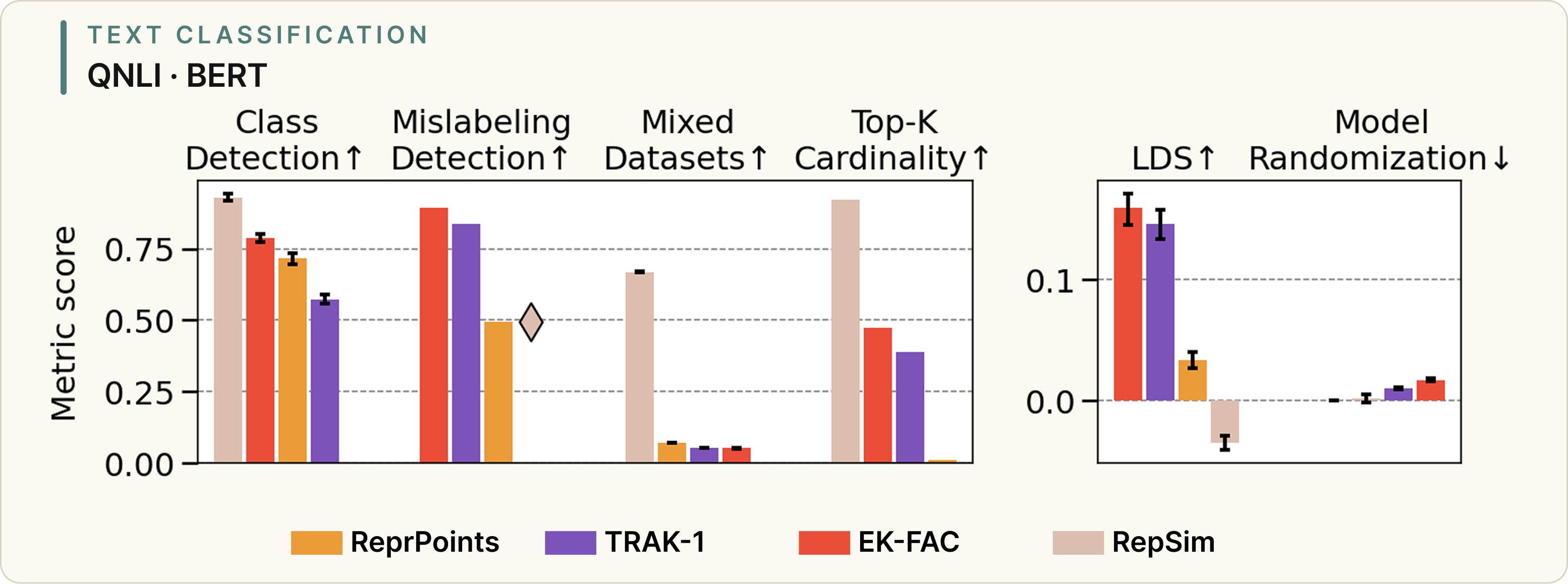}
\caption{Text-classification benchmarks (QNLI/BERT). Conventions as in~\Cref{fig:eval-image-classification}. The random baseline (mean $\pm$ std; $41$ seeds): $0.502{\pm}0.007$ (Class), $0.493{\pm}0.002$ (Mislabeling), $0.056{\pm}0.000$ (Mixed Datasets), $0.951{\pm}0.003$ (Top-$k$), $0.000{\pm}0.000$ (Model Rand.), and $0.000{\pm}0.003$ (LDS).}
\label{fig:eval-text-classification}
\end{figure*}

\begin{wrapfigure}{r}{0.4590896226 \linewidth}
    \centering
    \includegraphics[width=\linewidth]{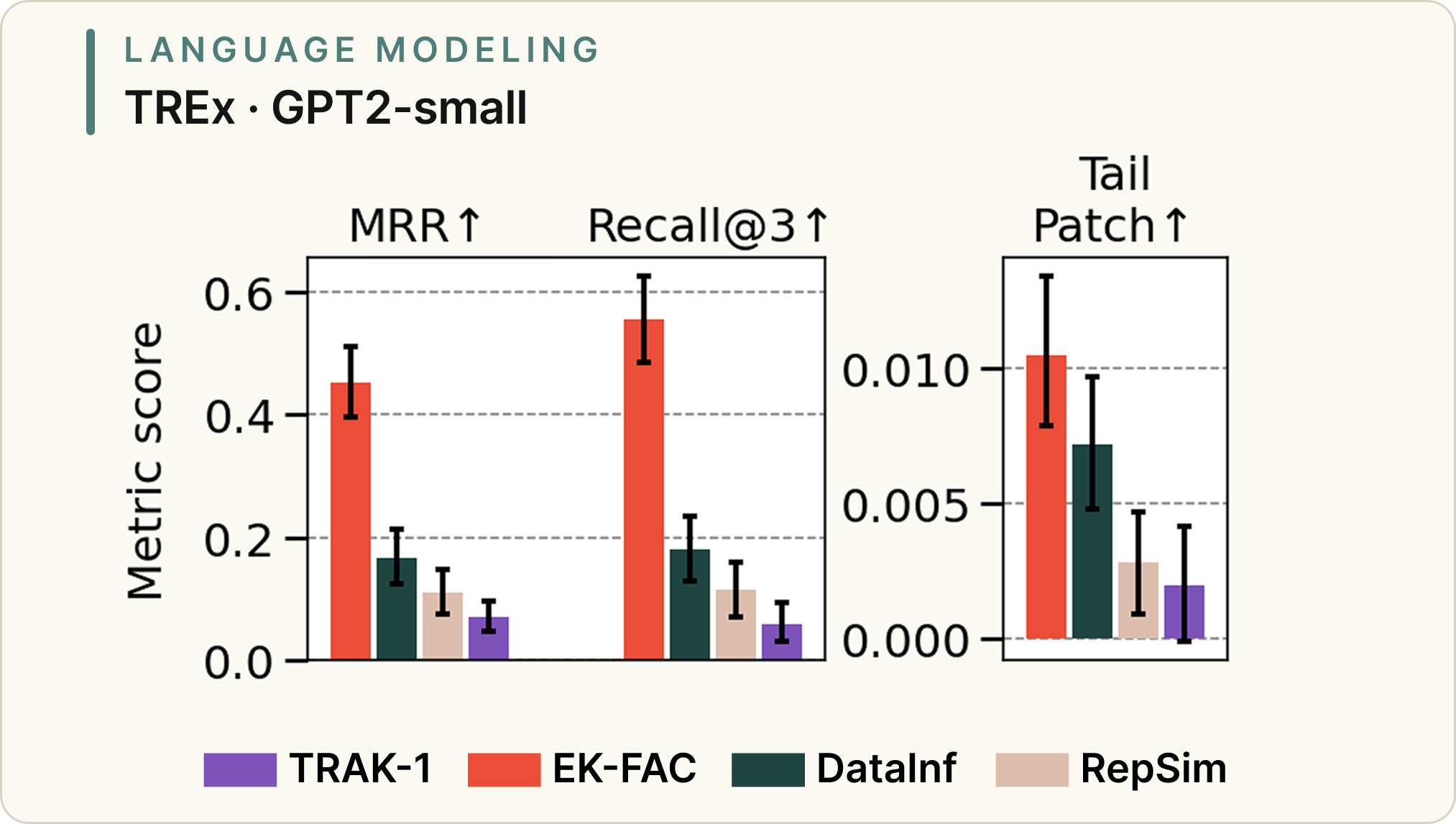}
    \caption{Language-modeling fact-tracing benchmarks (T-REx/GPT-2). Conventions as in~\Cref{fig:eval-image-classification}. Random baselines (mean $\pm$ std over $41$ seeds): $0.026{\pm}0.007$ (MRR), $0.014{\pm}0.009$ (Recall@$3$), and $-0.001{\pm}0.001$ (TailPatch).}
    \label{fig:eval-fact-tracing}
    \vskip -0.1in
\end{wrapfigure}

\Cref{fig:eval-image-classification,fig:eval-text-classification,fig:eval-fact-tracing} summarize our results across the three settings. Rankings shift substantially across metrics, exposing trade-offs between performance in different evaluation categories. In each figure, the left pane contains metrics with range $[0,1]$ and the right pane contains correlation-based metrics with range $[-1, 1]$. Results for differing benchmark and TDA hyperparameter values, including benchmark robustness checks, are reported in~\Cref{bench_robustness,tda_hyperparam}.

\paragraph{Image classification (AwA2 / ResNet-50)}
No method dominates across the metrics. TracInCP's relatively strong performance on the \emph{ground-truth} LDS metric does not translate to consistent \emph{downstream-task} performance, as it remains close to the random baseline on Subclass and Shortcut Detection. Still, the method's strong overall position suggests that aggregating gradient information across the training trajectory captures important signal that single-checkpoint methods miss. For \emph{downstream tasks} Subclass and Shortcut Detection, the cheap representation-space similarity baseline (RepSim) outperforms the more expensive TDA methods. This is unsurprising, given that these tasks reward identifying training samples that resemble the test example. The exception is Mislabeling Detection, where RepSim's score trivially collapses to $0.5$.
EK-FAC performs relatively well on Class and Mislabeling Detection but misses the mark on the other metrics, which we attribute in part to the reduced subset of layers used in the method due to computational constraints. Representer Points (ReprPoints) demonstrates mixed performance across tasks, consistent with the argument of~\citet{tsai2023sample} that its per-sample importance scores derive from loss gradients on the trained model and carry little signal once training has converged.
LDS scores are uniformly rather low, in line with prior reports~\citep{yolcu2025sparse}. TRAK-1, applied with a single checkpoint rather than the ensemble it was designed around, underperforms on most metrics, consistent with prior reports~\citep{park2023trak}. 

\paragraph{Text classification (QNLI / BERT)} 
The text classification results echo many patterns from the image classification benchmark.  RepSim dominates Class Detection and Mixed Datasets, where attributions that retrieve training points similar to the test example have an advantage. In contrast, the more modern gradient-based methods designed with counterfactual notions of TDA in mind, TRAK-1 and EK-FAC, dominate LDS, with both substantially outperforming RepSim. Apart from Class Detection, ReprPoints delivers near-random performance, which we again attribute to its sensitivity to model convergence. As in the image-classification setting, the strongest LDS performers are not the strongest performers across all \emph{downstream tasks}, reinforcing the importance of evaluating TDA methods across complementary criteria.

\vspace{2em}

\paragraph{Fact tracing (T-REx / GPT-2-small)}
In contrast to the classification settings, influence-function-based methods clearly prevail on fact tracing. EK-FAC leads on all three metrics, followed by DataInf, with RepSim and TRAK-1 trailing behind. EK-FAC's strength is consistent with its motivation as an influence-function surrogate designed to scale to language models~\citep{grosse2023studyinglargelanguagemodel}. The gap between EK-FAC and DataInf is unsurprising given DataInf's coarser approximation of the inverse Hessian. The low performance of TRAK-1 is again attributable to the method being designed to operate in an ensemble \citep{park2023trak}, and \citet{chang2025scalable} report similarly low TRAK results in their fact-tracing study.
All metrics agree on the method ordering in this setting, supporting tail-patch as a scalable proxy when entailing annotations are unavailable. Overall performance, however, remains poor, as at these metric scores, reliably retrieving fact-supporting training samples is effectively out of reach.

\section{Conclusion} \label{conclusion}
TDA methods have drawn growing interest in interpretability research, but their adoption has been held back by a lack of evaluation tooling. \quanda fills this gap with an open-source \texttt{Python} library unifying TDA evaluation, offering a broad set of metrics, ready-to-use benchmarks across image classification, text classification, and language modeling, and consistent wrappers around TDA implementations. Our empirical study shows that no method excels across all modalities and metrics. The best TDA approach depends on the intended application, underscoring the need for evaluation across multiple complementary axes.

\paragraph{Limitations and ongoing work.} Attribution is currently restricted to a single trained model, with no support for ensembled variants such as full TRAK. Language modeling evaluation is currently limited to fact tracing; per-configuration benchmark training for this modality is not yet supported, as it would require LLM pretraining at scale. We plan to address these in future releases alongside continuing extensions to \quanda's TDA wrappers, metrics, and benchmarks, and welcome community contributions through the public repository. Additionally, the evaluation protocols themselves carry limitations. Metrics such as Subclass, Mislabeling, and Shortcut Detection, Mixed Datasets Separation, and Fact Tracing require controlled settings and are not general-purpose quality estimators. \emph{Heuristics} are diagnostic metrics: a poor score signals a failure mode, but a good score does not imply high attribution quality. LDS applies to any setting but requires costly subset retraining; we recommend using the pre-trained benchmarks to avoid this. Finally, metrics can be sensitive to hyperparameters: in our analyses (\Cref{bench_robustness,tda_hyperparam}), rankings were largely stable, but absolute scores shifted considerably.

By lowering the cost of rigorous comparison, \quanda equips the community to evaluate TDA methods consistently and to ground future methodological progress in shared, reproducible benchmarks.

\begin{ack}

This work was supported by
the Fraunhofer Internal Programs as grant PREPARE (40-08394);
the Federal Ministry of Education and Research (BMBF) as grant BIFOLD (01IS18025A, 01IS180371I);
the German Research Foundation (DFG) as research unit DeSBi (KI-FOR 5363);
the European Union's Horizon Europe research and innovation programme (EU Horizon Europe) as grant TEMA (101093003);
the European Union's Horizon 2020 research and innovation programme (EU Horizon 2020) as grant iToBoS (965221);
and the European Union's Horizon Europe research and innovation programme (EU Horizon Europe) as grant ACHILLES (101189689).
Anna Hedstr\"{o}m is supported by the ETH AI Center Fellowship. The authors declare no competing interests.

\end{ack}

\vskip 0.2in
\bibliography{main}

\newpage
\appendix
\section{Notation} \label{notation}

We consider an ordered training dataset of size $n$, denoted as $\mathcal{D} = (\mathsf{z}_1, \dots, \mathsf{z}_n) \in \mathcal{Z}^n $, where each training sample $\mathsf{z}_i \in \mathcal{Z}$ represents an individual datapoint. In the context of classification tasks, a datapoint $\mathsf{z}_i$ consists of a pair \mbox{$\mathsf{z}_i = (\bm{x}_i, y_i)$}, where $\bm{x}_i \in \mathcal{X}$ is the input sample and $y_i \in [C]$ is the corresponding label. Here, $C \in \mathbb{N}$ denotes the number of classes (or vocabulary size), and, for any $m \in \mathbb{N}$, we write $[m]$ for the set $\{1, 2, \dots, m\}$. The input space $\mathcal{X}$ is $\mathbb{R}^d$ for image classification and a space of token sequences for text classification and language modelling. For test samples $\mathsf{z}=(\bm{x},y)\in \mathcal{Z}$, the label $y$ indicates the output of the network that was used for attributions, unless specified otherwise. 

The model we aim to explain is assumed to be a trained neural network, which implements the function $f(~\cdot~;\theta)$, where $\theta$ denotes the set of learned parameters of the model. We write $\theta^*(\mathcal{D}')$ for the parameters obtained by training on a dataset $\mathcal{D}' \subseteq \mathcal{D}$, and set $\theta := \theta^*(\mathcal{D})$. We denote the empirical risk:
\begin{equation*}\label{eq:risk}
\mathcal{J}(\theta;\mathcal{D})=\frac{1}{n}\sum_{i=1}^n \mathcal{L}(\mathsf{z}_i; \theta),
\end{equation*}
where $\mathcal{L}(\cdot\,;\cdot)$ is a sample-wise loss function, such as the cross-entropy loss.

We denote a function corresponding to the computational subgraph of the network up to an intermediate layer, i.e., a feature extractor, $h(~\cdot~; \vartheta): \mathcal{Z} \rightarrow \mathbb{R}^{d_h}$, with parameters $\vartheta$. For brevity, we will denote the hidden features as $h(\mathsf{z})$ and the model output as $f(\mathsf{z})$ in the following sections, and write $f_y(\mathsf{z}):= [f(\mathsf{z})]_y$ for the scalar output associated with the label $y$. 
The attribution for a test sample $\mathsf{z}$, computed by a TDA method, is denoted as $\tau(\mathsf{z}, \mathcal{D})_i$ for the training sample $\mathsf{z}_i$. We write $\mathcal{D}_{\text{test}}$ for the evaluation set, and $\pi^\tau_{\mathsf{z}}$ for the permutation that sorts the training samples in descending order of $\tau(\mathsf{z}, \mathcal{D})$. The remainder of the appendix uses this notation throughout.

\section{Training Data Attribution Methods} \label{tda_appendix}

This section provides theoretical definitions of the TDA methods evaluated in~\Cref{benchmarking} and~\Cref{app:extra_benchmark_results}, all of which are available through \appquanda. For the full list of TDA implementations wrapped by \appquanda, we refer the reader to the library documentation.

\subsection{Similarity of Representations}\label{appendix_similarity}
Gaining insights into what makes two samples similar from a model's perspective can aid practitioners in understanding the model's underlying reasoning and potentially detecting unwanted behaviours~\citep{charpiat2019input, hanawa2021evaluation}. The similarity between hidden representations of a given sample and those of training samples can be framed as an attribution method~\citep{caruana_case-based_1999}:

\begin{equation}
    \tau_{\text{SIM}}(\mathsf{z},\mathcal{D})_i=\sigma(h(\mathsf{z}),h(\mathsf{z}_i)),
\end{equation}

where $\sigma(\cdot,\cdot)$ is a similarity measure such as the dot product or the cosine similarity. For this method, $h$ may also be taken at an intermediate layer other than the penultimate one.

This explanation method is implemented as SimilarityInfluence in \texttt{Captum} \footnote{\href{https://github.com/pytorch/captum}{https://github.com/pytorch/captum}}. In \appquanda, we provide a wrapper around this \texttt{Captum} implementation. We refer to this method as RepSim throughout the paper.

\subsection{Influence Functions}\label{appendix_if}
Influence Functions (IF)~\citep{koh2017understanding} are a TDA method that approximates the counterfactual effect of retraining the model after removing a single training sample from the dataset. 
The approximation is computed by performing a single Newton optimization step on the counterfactual loss landscape, which results in the following attribution:
\begin{equation}\label{eq:IF}
\tau_{\text{IF}}(\mathsf{z},\mathcal{D})_i =\nabla_{\theta}\mathcal{L}(\mathsf{z};\theta)^\top H^{-1}_{\theta}\nabla_\theta \mathcal{L}(\mathsf{z}_i;\theta),
\end{equation}
where $H_\theta=\nabla^2_\theta\mathcal{J}(\theta;\mathcal{D})=\frac{1}{n}\sum_{i=1}^n\nabla^2_\theta\mathcal{L}(\mathsf{z}_i;\theta)$ is the Hessian.

Computing the inverse Hessian is a computationally demanding task, and the approach introduced in the original paper~\citep{koh2017understanding} incurs a substantial memory footprint. To mitigate these challenges, recent work~\citep{schioppa2022scaling} has proposed to speed up the inverse Hessian calculation using Arnoldi iterations~\citep{arnoldi1951principle}. This method is named ArnoldiInfluenceFunction in \texttt{Captum}. \appquanda provides a wrapper around this implementation.

To generate a global ranking of training samples, such as for detecting mislabeled examples, \citet{koh2017understanding} propose evaluating the influence of each sample on itself, following~\Cref{eq:IF}. In \appquanda, we also provide a wrapper around the \texttt{Captum} implementation to compute these self-influences.

\subsection{TracIn}\label{appendix_tracin}
TracIn~\citep{pruthi2020estimating} builds on the idea of tracking how the loss on a test point evolves throughout the training process. The change in the loss for a test sample $\mathsf{z}$ caused by a training sample $\mathsf{z}_i$ is approximated via a linear approximation:
\begin{equation}
\label{eq:TracIn}
    \tau_{\text{TracIn}}(\mathsf{z},\mathcal{D})_i=\sum_t \eta_t \nabla_{\theta_t}\mathcal{L}(\mathsf{z};\theta_t)^\top \nabla_{\theta_t}\mathcal{L}(\mathsf{z}_i; \theta_t),
\end{equation}
where $t$ indexes the training iterations in which the training sample $\mathsf{z}_i$ was used, $\theta_t$ represents the parameters and $\eta_t$ the learning rate at iteration $t$.

The full TracIn attributor as defined in~\Cref{eq:TracIn} is prohibitively computationally expensive, as it requires storing the parameters at each training step. To address this, attributions are computed using only a selected set of checkpoints, a method known as TracInCP. \texttt{Captum} implements this approach with two separate extensions, as described in the original paper~\citep{pruthi2020estimating}:
\begin{itemize}
    \item TracInCPFast: This variant simplifies the process further by considering only the final layer parameters when computing the gradients.
    \item TracInCPFastRandProj: This version additionally uses random projections to reduce the dimensionality of the gradients, thereby speeding up the computations.
\end{itemize}

\appquanda provides wrappers for TracInCP and both extensions.

\subsection{Representer Points}\label{appendix_representer}
The Representer Points (ReprPoints) method~\citep{yeh2018representer} leverages a representer theorem for deep neural networks. Consider a feature extractor $h_p$, given by the penultimate layer, followed by a linear classifier with weight matrix $W \in \mathbb{R}^{d_h \times C}$ and bias $\bm{b} \in \mathbb{R}^C$. We can express the model output as $f(\mathsf{z}; \theta) = W^\top h_p(\mathsf{z}; \vartheta) + \bm{b}$. Under the assumption that the model is trained to convergence with $L_2$ regularization, the final layer parameters $W$ can be written as a linear combination of final layer features $h_p(\mathsf{z}_i)$. This formulation reduces the model to a kernel machine, enabling a decomposition of the model output into the contributions of each training point. This results in the following attribution function:
\begin{equation}
    \label{eq:RP}
    \tau_{\text{ReprPoints}}(\mathsf{z},\mathcal{D})_i=-\left[\frac{\partial\mathcal{L}(\mathsf{z}_i;\theta)}{\partial f(\mathsf{z}_i)}\right]_y\,h_p(\mathsf{z}_i)^\top h_p(\mathsf{z}),
\end{equation}

Given a model that is not originally trained with $L_2$ regularization, the Representer Points approach suggests training the final layer parameters $W$ with $L_2$ regularization, using backtracking line search to ensure convergence.

\appquanda provides a wrapper around the official code release by the original paper's authors\footnote{\href{https://github.com/chihkuanyeh/Representer_Point_Selection}{https://github.com/chihkuanyeh/Representer\_Point\_Selection}}. 

\subsection{TRAK}\label{appendix_trak}
TRAK~\citep{park2023trak} also leverages the leave-one-out perspective, similar to~\Cref{appendix_if} but, rather than estimating the inverse Hessian directly, approximates the counterfactual effect of retraining on a subset of training points by linearizing the model around the model parameters. The first-order Taylor approximation of the model output corresponds to using the empirical Neural Tangent Kernel~\citep{jacot2018neural} as a surrogate for the model.

TRAK is defined to use multiple independent instantiations of the model, trained on the same dataset or different subsets of the dataset. Since \appquanda currently considers post-hoc attribution of a single model, we consider the case of using only a single model, whose behavior we want to attribute to the training data, which we refer to as TRAK-1.

Similar to TracIn, TRAK employs a random projection $P\in\mathbb{R}^{|\theta|\times k}$, $k\ll|\theta|$, on the gradients. Let $\phi (\mathsf{z}) = P^\top\nabla_\theta f_y(\mathsf{z})$ be the gradient of the model output, where we define $\phi_i = \phi (\mathsf{z}_i)$ for a training point $\mathsf{z}_i$. Let $\Phi=\bigl[\phi_1^\top;\phi_2^\top;\dots;\phi_n^\top\bigr]\in\mathbb{R}^{n\times k}$ be the matrix of projected gradients. Additionally, let $Q$ be the $n \times n$ diagonal matrix such that $Q_{i,i}=1-p(\mathsf{z}_i;\theta)$, where $p(\mathsf{z};\theta)$ denotes the probability of the label $y$ of $\mathsf{z}$. In the binary classification scenario, TRAK-1 can then be formulated as follows:
\begin{equation}
    \label{eq:TRAK}
    \tau_{\text{TRAK}}(\mathsf{z},\mathcal{D})_i=\bigl[\phi(\mathsf{z})^\top (\Phi^\top \Phi)^{-1}\Phi^\top Q\bigr]_i.
\end{equation}

To extend the formulation to the multiclass case, we replace the model output $f_y(\cdot)$ with the function $r(\mathsf{z})=\log\Bigl(\frac{p(\mathsf{z};\theta)}{1-p(\mathsf{z};\theta)}\Bigr)$. For test points, this function represents the margin of the output that is chosen for explanation.

TRAK offers an official code release\footnote{\href{https://github.com/MadryLab/trak}{https://github.com/MadryLab/trak}}, which can be conveniently used with its \appquanda wrapper.

\subsection{EK-FAC Influence Functions}\label{appendix_ekfac}
EK-FAC influence functions~\citep{grosse2023studyinglargelanguagemodel} address the central computational bottleneck of~\Cref{eq:IF} by replacing the inverse Hessian with a tractable curvature surrogate built from the Gauss--Newton Hessian $G$, which for the cross-entropy loss equals the Fisher Information Matrix (FIM). Building on the simplifying assumption that second-order interactions of parameters in different layers are negligible, K-FAC~\citep{martens2015kfac} reduces $G$ to a block-diagonal matrix with a block $G_\ell$ for each layer. For a layer with input activations $\bm{a}_{\ell-1}$ and pre-activation pseudo-gradients $\bm{s}_\ell$, computed with labels sampled from the model's output distribution rather than taken from the dataset, K-FAC further assumes the two are independent and approximates the block by the Kronecker product
\begin{equation}
    G_\ell\approx\mathbb{E}[\bm{a}_{\ell-1}\bm{a}_{\ell-1}^\top]\,\otimes\,\mathbb{E}[\bm{s}_\ell\bm{s}_\ell^\top]=A_{\ell-1}\otimes S_\ell,
\end{equation}
which admits an analytic inverse $A_{\ell-1}^{-1}\otimes S_\ell^{-1}$. EK-FAC~\citep{george2018ekfac} additionally keeps the eigenbasis of $A_{\ell-1}\otimes S_\ell$ but replaces its eigenvalues with the second moments of the weight pseudo-gradients projected onto this basis, yielding $\hat{G}_\ell$. \citet{grosse2023studyinglargelanguagemodel} show that this surrogate scales influence-function-style attribution to large language models. Plugging it into~\Cref{eq:IF} layer by layer yields the EK-FAC attribution
\begin{equation}
    \tau_{\text{EK-FAC}}(\mathsf{z},\mathcal{D})_i=\sum_{\ell}\nabla_{\theta_\ell}\mathcal{L}(\mathsf{z};\theta)^\top\,\bigl(\hat{G}_\ell+\lambda I\bigr)^{-1}\,\nabla_{\theta_\ell}\mathcal{L}(\mathsf{z}_i;\theta),
\end{equation}
where $\theta_\ell$ are the parameters of layer $\ell$ and $\lambda>0$ is a damping term. \appquanda exposes EK-FAC through two wrappers: one around the \texttt{kronfluence} package\footnote{\href{https://github.com/pomonam/kronfluence}{https://github.com/pomonam/kronfluence}} and one around the EK-FAC attributor in \texttt{dattri}~\citep{dattri}\footnote{\href{https://github.com/TRAIS-Lab/dattri}{https://github.com/TRAIS-Lab/dattri}}.

\subsection{DataInf}\label{appendix_datainf}
DataInf~\citep{kwon2024datainf}, similar to EK-FAC, also replaces the inverse Hessian in~\Cref{eq:IF} with a tractable per-layer surrogate, but designs a closed-form approximation of the inverse Hessian-vector product that avoids any matrix inversion or iterative solver. Following EK-FAC's simplifying assumption that second-order effects are negligible between layers, DataInf treats the Hessian per layer and replaces $H_{\theta_\ell}$ with the (damped) empirical Fisher,
\begin{equation}
    \widehat{F}_\ell\;=\;\frac{1}{n}\sum_{j=1}^{n}g_{\ell,j}g_{\ell,j}^\top + \lambda I,
\end{equation}
where $g_{\ell,j}=\nabla_{\theta_\ell}\mathcal{L}(\mathsf{z}_j;\theta)$ is the gradient of the loss and $\lambda>0$ is a damping coefficient. Then DataInf proposes swapping the order of summation and matrix inversion:

\begin{equation}
        \widehat{F}_\ell^{-1}\;=\;\left(\frac{1}{n}\sum_{j=1}^{n}g_{\ell,j}g_{\ell,j}^\top + \lambda I\right)^{-1}\;\approx\;\frac{1}{n}\sum_{j=1}^{n}\left(g_{\ell,j}g_{\ell,j}^\top + \lambda I\right)^{-1}.
\end{equation}

Finally, DataInf applies the Sherman--Morrison identity in order to reach a closed-form solution with no matrix inversions. This yields the per-layer inverse-Hessian approximation
\begin{equation}
    \widehat{F}_\ell^{-1}\;\approx\;\frac{1}{n\lambda} \sum_{j=1}^{n} \!\left(I-\frac{g_{\ell,j}g_{\ell,j}^\top}{\lambda+\lVert g_{\ell,j}\rVert^{2}}\,\right).
\end{equation}
The attribution then reads
\begin{equation}
    \tau_{\text{DataInf}}(\mathsf{z},\mathcal{D})_i=\sum_{\ell}\nabla_{\theta_\ell}\mathcal{L}(\mathsf{z};\theta)^\top\,\widehat{F}_\ell^{-1}\,\nabla_{\theta_\ell}\mathcal{L}(\mathsf{z}_i;\theta).
\end{equation}
\appquanda wraps the DataInf attributor from \texttt{dattri}, which exposes the layer set, damping ($\lambda$), and a sub-sampling ratio for the empirical Fisher estimate.

\section{Benchmarks} \label{benchmarks_appendix}
Each \texttt{Metric} described in~\Cref{metrics_desc} has its own class in \appquanda and is paired with a corresponding \texttt{Benchmark} class. The role of these classes is to lift the burden of constructing the controlled environment that the metric requires: while a \texttt{Metric} expects the caller to supply all ingredients, including a (possibly transformed) training dataset, a model trained on it, and the attribution scores, a \texttt{Benchmark} manages dataset manipulation, model training, and the generation of explanations, then delegates scoring to the underlying \texttt{Metric}.

Since benchmark construction can itself be tricky, \appquanda exposes a library of ready-made benchmarks via the \texttt{load\_pretrained} initializer method, providing a standardized evaluation environment that researchers can compare against without rerunning the training pipeline. The benchmark assets are currently hosted at \url{https://huggingface.co/quanda-bench-test}. We envision that new benchmark settings can be prepared by tweaking the parameters of existing configurations rather than writing them from scratch. Training procedures and properties of the released benchmarks are described in~\Cref{exp_details}.

\section{Experimental Details} \label{exp_details}

In this section, we document the configurations of the benchmarks released within \appquanda, as well as the details of the experiments presented in~\Cref{benchmarking}. All the described settings, along with the codenames used to refer to them in the rest of this section, are listed in~\Cref{tab:bench-codenames}. The remainder of the
Appendix is organized into five subsections: the datasets used for
each benchmark (\Cref{exp-datasets}), the training recipes for the
source models on top of which the benchmarks are evaluated
(\Cref{exp-training}), the metric- and benchmark-level configuration
values (\Cref{exp-bench-metrics}), the details about the fact-tracing
benchmarks (\Cref{fact-tracing-details}), and the details about the TDA
methods used (\Cref{exp-explainers}).

\subsection{Datasets}\label{exp-datasets}

For each setting, the dataset splits are reported in~\Cref{tab:bench-datasets}, and the adversarial second source merged into label $0$ for every Mixed Datasets benchmark is listed in~\Cref{tab:bench-adv}. The source and license of every asset loaded by the ready-made benchmark configurations are presented in~\Cref{tab:hf-assets}. Finally, the benchmark-specific transformations applied on top of the source dataset (when required by
the evaluation metric) are detailed in~\Cref{tab:bench-wrappers}.

\begin{table}[tb]
\centering
\small
\caption{Benchmark settings used throughout~\Cref{exp_details}. Each
(model, dataset) combination is referred to by the codename in the
first column. The \textbf{QNLI} suite initializes from
\texttt{bert-base-cased} pretrained weights (see~\Cref{tab:hf-assets}), all other suites are
trained from scratch. We modify the ResNet-9 and BERT models as described in the TRAK paper~\citep{park2023trak}.}
\label{tab:bench-codenames}
\begin{tabular}{@{}lll@{}}
\toprule
Codename & Model & Dataset \\
\midrule
\textbf{MNIST} & LeNet-5~\citep{lecun_1998_lenet}        & MNIST~\citep{lecun1998mnist}              \\
\textbf{CIFAR} & ResNet-9~\citep{he_2016_cvpr}          & CIFAR-10~\citep{krizhevsky2009learning}   \\
\textbf{AwA2}  & ResNet-50~\citep{he_2016_cvpr}           & AwA2~\citep{xian2018zero}                 \\
\textbf{QNLI}  & fine-tuned BERT~\citep{devlin-etal-2019-bert} & QNLI~\citep{wang-etal-2018-glue} (SetFit, $50$k) \\
\textbf{T-REx} & GPT-2 small~\citep{radford2019gpt2}    & \makecell[l]{T-REx~\citep{elsahar-etal-2018-rex} \\ + $10\%$ OpenWebText~\citep{Gokaslan2019OpenWeb}} \\
\bottomrule
\end{tabular}
\end{table}

\begin{table}[tb]
\centering
\small
\caption{Train/validation/evaluation splits used by the classification benchmarks. Validation splits are used for reporting model performance. Evaluation datasets are used for calculating metric results.
Rows marked $\hookrightarrow$ list the adversarial second source
merged into Mixed Datasets dataset splits as a single class (label $0$);
their ratios are fractions of the adversarial dataset's own train split,
not of the base dataset.
See~\Cref{tab:bench-codenames} for codenames.}
\label{tab:bench-datasets}
\begin{tabular}{@{}llc@{}}
\toprule
Setting & Benchmark & Train/Val/Test \\
\midrule
\multirow{7}{*}{\textbf{MNIST}}
 & Class Detection                                  & $0.80\,\text{Train}\,/\,0.20\,\text{Train}\,/\,\text{Test}$ \\
 & Subclass Detection                               & $0.80\,\text{Train}\,/\,0.20\,\text{Train}\,/\,\text{Test}$ \\
 & Mislabeling Detection                            & $0.80\,\text{Train}\,/\,0.20\,\text{Train}\,/\,\text{Test}$ \\
 & Shortcut Detection                               & $0.80\,\text{Train}\,/\,0.20\,\text{Train}\,/\,\text{Test}$ \\
 & LDS                                             & $0.80\,\text{Train}\,/\,0.20\,\text{Train}\,/\,\text{Test}$ \\
 & Mixed Datasets                                   & $0.80\,\text{Train}\,/\,0.20\,\text{Train}\,/\,\text{Test}$ \\
 & \quad$\hookrightarrow$ Fashion-MNIST adv.       & $0.80\,\text{Train}\,/\,0.10\,\text{Train}\,/\,0.10\,\text{Train}$ \\
\cmidrule(lr){1-3}
\multirow{7}{*}{\textbf{CIFAR}}
 & Class Detection                                  & $0.80\,\text{Train}\,/\,0.20\,\text{Train}\,/\,\text{Test}$ \\
 & Subclass Detection                               & $0.80\,\text{Train}\,/\,0.20\,\text{Train}\,/\,\text{Test}$ \\
 & Mislabeling Detection                            & $0.80\,\text{Train}\,/\,0.20\,\text{Train}\,/\,\text{Test}$ \\
 & Shortcut Detection                               & $0.80\,\text{Train}\,/\,0.20\,\text{Train}\,/\,\text{Test}$ \\
 & LDS                                             & $0.80\,\text{Train}\,/\,0.20\,\text{Train}\,/\,\text{Test}$ \\
 & Mixed Datasets                                   & $0.80\,\text{Train}\,/\,0.20\,\text{Train}\,/\,\text{Test}$ \\
 & \quad$\hookrightarrow$ SVHN (cropped) adv.      & $0.16\,\text{Train}\,/\,0.02\,\text{Train}\,/\,0.02\,\text{Train}$ \\
\cmidrule(lr){1-3}
\multirow{7}{*}{\textbf{AwA2}}
 & Class Detection                                  & $0.80\,\text{Train}\,/\,0.1\,\text{Train}\,/\,0.1\,\text{Train}$ \\
 & Subclass Detection                               & $0.80\,\text{Train}\,/\,0.1\,\text{Train}\,/\,0.1\,\text{Train}$ \\
 & Mislabeling Detection                            & $0.80\,\text{Train}\,/\,0.1\,\text{Train}\,/\,0.1\,\text{Train}$ \\
 & Shortcut Detection                               & $0.80\,\text{Train}\,/\,0.1\,\text{Train}\,/\,0.1\,\text{Train}$ \\
 & LDS                                             & $0.80\,\text{Train}\,/\,0.1\,\text{Train}\,/\,0.1\,\text{Train}$ \\
 & Mixed Datasets                                   & $0.80\,\text{Train}\,/\,0.1\,\text{Train}\,/\,0.1\,\text{Train}$ \\
 & \quad$\hookrightarrow$ ImageNet-Sketch adv.     & $0.01\,\text{Train}\,/\,0.002\,\text{Train}\,/\,0.1\,\text{Train}$ \\
\cmidrule(lr){1-3}
\multirow{5}{*}{\textbf{QNLI}}
 & Class Detection                                  & $0.48\,\text{Train}\,/\,0.1\,\text{Train}\,/\,\text{Test}$ \\
 & Mislabeling Detection                            & $0.48\,\text{Train}\,/\,0.1\,\text{Train}\,/\,\text{Test}$ \\
 & LDS                                             & $0.48\,\text{Train}\,/\,0.1\,\text{Train}\,/\,\text{Test}$ \\
 & Mixed Datasets                                   & $0.48\,\text{Train}\,/\,0.1\,\text{Train}\,/\,\text{Test}$ \\
 & \quad$\hookrightarrow$ MRPC (SetFit) adv.       & $0.80\,\text{Train}\,/\,0.1\,\text{Train}\,/\,0.1\,\text{Train}$ \\
\bottomrule
\end{tabular}
\end{table}

\begin{table}[tb]
\centering
\small
\caption{Adversarial dataset source used in Mixed Datasets training.}
\label{tab:bench-adv}
\begin{tabular}{@{}lll@{}}
\toprule
Setting & Adversarial dataset & Citation \\
\midrule
\textbf{MNIST} & Fashion-MNIST   & \citet{xiao2017online}   \\
\textbf{CIFAR} & SVHN (cropped)  & \citet{netzer2011reading}         \\
\textbf{AwA2}  & ImageNet-Sketch & \citet{wang2019learning} \\
\textbf{QNLI}  & MRPC  (GLUE)           & \citet{wang-etal-2018-glue}          \\
\bottomrule
\end{tabular}
\end{table}

\begin{table}[tb]
\centering
\small
\caption{Sources and licenses of assets loaded by ready-made benchmark
configurations, grouped by setting. \texttt{HF:} denotes \texttt{Hugging Face} Hub repos.}
\label{tab:hf-assets}
\begin{tabularx}{\columnwidth}{@{}l l X l@{}}
\toprule
Setting & Asset & Source & License \\
\midrule
\multirow{2}{*}{\textbf{MNIST}}
 & Train/Val/Test      & \texttt{ylecun/mnist}                    & MIT \\
 & Mixed Datasets adv.  & \texttt{zalando-datasets/fashion\_mnist} & MIT \\
\cmidrule(lr){1-4}
\multirow{2}{*}{\textbf{CIFAR}}
 & Train/Val/Test      & \texttt{uoft-cs/cifar10}                              & MIT \\
 & Mixed Datasets adv.  & \texttt{ufldl-stanford/svhn} (\texttt{cropped\_digits}) & Non-comm. \\
\cmidrule(lr){1-4}
\multirow{2}{*}{\textbf{AwA2}}
 & Train/Val/Test      & \url{cvml.ista.ac.at/AwA2}              & Research \\
 & Mixed Datasets adv.  & \texttt{Huanyiiiii/Imagenet\_Sketch}    & MIT \\
\cmidrule(lr){1-4}
\multirow{3}{*}{\textbf{QNLI}}
 & Model + tokenizer   & \texttt{google-bert/bert-base-cased} & Apache-2.0 \\
 & Train/Val/Test      & \texttt{SetFit/qnli}                 & CC BY-SA 4.0 \\
 & Mixed Datasets adv.  & \texttt{SetFit/mrpc}                 & Non-comm. \\
\cmidrule(lr){1-4}
\textbf{T-REx}
 & Prompts / evidence  & \url{github.com/PAIR-code/pretraining-tda} & CC BY-SA 4.0 \\
\bottomrule
\end{tabularx}
\end{table}

\begin{table}[tb]
\centering
\small
\caption{Train- and evaluation-set transformations applied to construct
each benchmark. Sweep grids are shown in $\{\cdot\}$, with the
selected value underlined. See~\Cref{tab:bench-codenames} for codenames.}
\label{tab:bench-wrappers}
\begin{tabularx}{\linewidth}{@{}ll>{\raggedright\arraybackslash}X>{\raggedright\arraybackslash}X@{}}
\toprule
Setting & Benchmark & Train transformation & Evaluation transformation \\
\midrule
\multirow{3}{*}{\textbf{MNIST}}
 & Subclass Detection
  & $10$ digit classes randomly partitioned into $2$ superclasses.
  & Same partition as training. \\
 & Mislabeling Detection
  & $\{\underline{10\%}, 20\%\}$ of training labels replaced with a
    uniformly random non-true class.
  & --- \\
 & Shortcut Detection
  & White $4{\times}4$ patch added to $\{75\%, 90\%, \underline{95\%}\}$
    of digit-$0$ training images.
  & White patch added to all images. \\
\cmidrule(lr){1-4}
\multirow{3}{*}{\textbf{CIFAR}}
 & Subclass Detection
  & $10$ classes partitioned into $2$ superclasses: animals and vehicles.
  & Same partition as training. \\
 & Mislabeling Detection
  & $\{\underline{10\%}, 20\%\}$ drawn i.i.d. across training samples relabeled with non-true class.
  & --- \\
 & Shortcut Detection
  & Yellow square added to $\{\underline{75\%}, 90\%, 95\%\}$ of
    \emph{airplane} training images.
  & Yellow square added to all
    images. \\
\cmidrule(lr){1-4}
\multirow{3}{*}{\textbf{AwA2}}
 & Subclass Detection
  & $50$ animal classes randomly partitioned into $2$ superclasses.
  & Same partition as training. \\
 & Mislabeling Detection
  & $10\%$ drawn i.i.d. across training samples relabeled with a non-true class.
  & --- \\
 & Shortcut Detection
  & Pink frame added to $85\%$ of class-$0$ training images.
  & Pink frame added to all images. \\
\cmidrule(lr){1-4}
\textbf{QNLI}
 & Mislabeling Detection
  & $2\%$ drawn i.i.d. across training samples swapped to the opposite class.
  & --- \\
\bottomrule
\end{tabularx}

\vspace{2pt}
\end{table}

\subsection{Benchmark Model Training}\label{exp-training}
Most benchmarks in our ready-made suite require models to be trained under benchmark-specific conditions. To determine the best benchmark-specific hyperparameters and training configuration, we performed a grid search over the combined set of these parameters. The benchmark-specific hyperparameters included in the grid search were the percentage of corrupted data in Mislabeling Detection and the percentage of samples in the selected class with the shortcut artifact inserted in Shortcut Detection. The search grid and the selected values are reported in~\Cref{tab:bench-wrappers}.

Optimizer, schedule, learning rate, and the number of epochs for every classification suite are summarized in~\Cref{tab:bench-trainer}, with search grid parameters shown in $\{\cdot\}$ and the selected setting underlined. In selecting the training (\Cref{tab:bench-trainer}) and benchmark hyperparameters (\Cref{tab:bench-wrappers}), our explicit objective was to achieve good performance across all the reported performance indicators, including validation accuracy and benchmark-specific indicators, such as high level of shortcut memorization for the Shortcut Detection metric, correct classification of adversarial data points for the Mixed Datasets benchmark, and the memorization of corrupted samples at strong validation accuracy for the Mislabeling Detection benchmark, as specified in the ``Additional Detail'' column of~\Cref{tab:bench-results}.

For Shortcut Detection, a single patch over the images is applied uniformly to a fraction of training images of a single class, in our pre-trained benchmark setting of class 0.~\Cref{tab:bench-results} reports the train and validation accuracies of the ready-made benchmarks available in \appquanda on which the evaluation in this paper is based, together with benchmark-specific sanity-check quantities used to verify that each benchmark model encodes desirable properties of the evaluation settings.

\begin{table}[tb]
\centering
\small
\caption{Training hyperparameters for the source models of each
benchmark suite. The BERT checkpoint is the only HF model loaded; all other source
models are trained from scratch. See~\Cref{tab:bench-codenames} for codenames.}
\label{tab:bench-trainer}
\begin{tabular}{@{}lllllr@{}}
\toprule
Setting & Benchmark & Optimizer & Sched. & Learning Rate & Epochs \\
\midrule
\multirow{6}{*}{\textbf{MNIST}}
 & Class Detection         & SGD                                       & const & $\{ 0.001, 0.01, \underline{0.1}\}$    & $15$ \\
 & Subclass Detection      & SGD                                       & const & $\{0.001, 0.01, \underline{0.1}\}$    & $15$ \\
 & Mislabeling Detection   & $\{\underline{\text{Adam}}, \text{SGD}\}$ & const & $\{0.0001, \underline{0.001}\}$       & $\{50, 100, \underline{200}\}$ \\
 & Shortcut Detection      & $\{\underline{\text{Adam}}, \text{SGD}\}$ & const & $\{0.0001, 0.001, \underline{0.01}\}$ & $\{5, \underline{50}, 100\}$   \\
 & LDS                    & Adam                                      & const & $\{0.001, \underline{0.01}\}$         & $5$ \\
 & Mixed Datasets          & SGD                                       & const & $\{0.001, 0.01, \underline{0.1}\}$    & $15$ \\
\cmidrule(lr){1-6}
\multirow{6}{*}{\textbf{CIFAR}}
 & Class Detection         & SGD  & const & $\{0.1, \underline{0.2}, 0.4\}$ & $\{12, 24, \underline{36}\}$         \\
 & Subclass Detection      & SGD  & const & $\{0.1, 0.2, \underline{0.4}\}$ & $\{12, 24, \underline{36}\}$         \\
 & Mislabeling Detection   & SGD  & const & $\{0.1, \underline{0.2}, 0.4\}$ & \makecell[r]{$\{12, 24, 36,$\\ $48, \underline{96}\}$} \\
 & Shortcut Detection      & SGD  & const & $\{0.1, \underline{0.2}, 0.4\}$ & \makecell[r]{$\{12, 24, 36,$\\ $48, \underline{96}\}$} \\
 & LDS                    & SGD  & const & $\{0.1, \underline{0.2}, 0.4\}$ & $\{12, 24, \underline{36}\}$         \\
 & Mixed Datasets          & SGD  & const & $\{\underline{0.1}, 0.2, 0.4\}$ & $\{48, \underline{96}\}$             \\
\cmidrule(lr){1-6}
\multirow{6}{*}{\textbf{AwA2}}
 & Class Detection         & SGD  & 1cycle & $\{0.03, 0.1, \underline{0.2}\}$ & $30$                       \\
 & Subclass Detection      & SGD  & 1cycle & $\{0.03, \underline{0.1}, 0.2\}$ & $30$                       \\
 & Mislabeling Detection   & SGD  & 1cycle & $\{0.03, 0.1, \underline{0.2}\}$ & $\{\underline{64}, 96\}$   \\
 & Shortcut Detection      & SGD  & 1cycle & $\{0.1, \underline{0.2}\}$       & $64$                       \\
 & LDS                    & SGD  & 1cycle & $\{0.03, 0.1, \underline{0.2}\}$ & $30$                       \\
 & Mixed Datasets          & SGD  & 1cycle & $\{0.1, \underline{0.2}\}$       & $\{\underline{30}, 96\}$   \\
\cmidrule(lr){1-6}
\multirow{4}{*}{\textbf{QNLI}}
 & Class Detection         & SGD   & linear & $10^{-3}$ & $20$ \\
 & Mislabeling Detection   & SGD   & linear & $10^{-3}$ & $30$ \\
 & LDS                    & SGD   & linear & $10^{-3}$ & $20$ \\
 & Mixed Datasets          & SGD   & linear & $10^{-3}$ & $20$ \\
\cmidrule(lr){1-6}
\multirow{2}{*}{\textbf{T-REx}}
& Fact-tracing (pretrain)  & AdamW & cosine & \makecell[l]{peak $6{\times}10^{-4}$,\\ min $6{\times}10^{-5}$} & $30$k it. \\
 & Fact-tracing (fine-tune) & AdamW & linear & $5{\times}10^{-5}$ & $3$        \\
\bottomrule
\end{tabular}

\end{table}

\begin{table}
\centering
\small
\caption{Train and validation accuracies of the source models for every
released benchmark, together with benchmark-specific diagnostics. ``eval shortcut success'' in Shortcut Detection refers to the share of the full evaluation set (1000 samples in our evaluations) that the model both mislabels and predicts as the target class (i.e., samples "swayed" by the shortcut; only such samples are used for metric calculation.)}
\label{tab:bench-results}
\begin{tabularx}{\linewidth}{@{}llcc>{\raggedright\arraybackslash}X@{}}
\toprule
Setting & Benchmark & \makecell{Train\\acc.} & \makecell{Val.\\acc.} & Additional Details \\
\midrule
\multirow{8}{*}{\textbf{MNIST}}
 & Class Detection       & 99.65\% & 98.67\% & --- \\
 & TopKCardinality      & 99.65\% & 98.67\% & --- \\
 & ModelRandomization   & 99.65\% & 98.67\% & --- \\
 & Subclass Detection    & 99.78\% & 99.18\% & --- \\
 & Mixed Datasets        & 99.89\% & 99.16\% & adversarial dataset\ train/eval 100.00\%/100.00\% \\
 & Mislabeling Detection & 98.28\% & 90.00\% & mislabeling memorization\ 88.55\% \\
 & Shortcut Detection    & 91.90\% & 87.67\% &  shortcut memorization train 99.82\%, eval shortcut success 19.25\% \\
 & LDS                  & 99.22\% & 98.65\% & 3 random subsets acc.\ 98.25\%/97.85\%/98.34\% \\
\cmidrule(lr){1-5}
\multirow{8}{*}{\textbf{CIFAR}}
 & Class Detection       & 100.00\% & 87.82\% & --- \\
 & TopKCardinality      & 100.00\% & 87.82\% & --- \\
 & ModelRandomization   & 100.00\% & 87.82\% & --- \\
 & Subclass Detection    & 100.00\% & 93.60\% & --- \\
 & Mixed Datasets        & 100.00\% & 88.74\% & adversarial dataset\ train/eval 100.00\%/99.73\% \\
 & Shortcut Detection    & 100.00\% & 87.45\% & shortcut memorization train 100.00\%, eval shortcut success 89.93\% \\
 & Mislabeling Detection & 100.00\% & 82.30\% & mislabeling memorization\ 100.00\% \\
 & LDS                  & 100.00\% & 87.82\% & 3 random subsets acc.\ 83.37\%/82.61\%/83.34\% \\
\cmidrule(lr){1-5}
\multirow{8}{*}{\textbf{AwA2}}
 & Class Detection       & 99.91\% & 80.23\% & --- \\
 & TopKCardinality      & 99.92\% & 80.23\% & --- \\
 & ModelRandomization   & 99.93\% & 80.23\% & --- \\
 & Subclass Detection    & 99.30\% & 72.43\% & --- \\
 & Mixed Datasets        & 99.91\% & 79.78\% & adversarial dataset\ train/eval 100.00\%/94.65\% \\
 & Shortcut Detection    & 99.91\% & 79.37\% & shortcut memorization train 100.00\%, eval shortcut success 93.91\% \\
 & Mislabeling Detection & 99.92\% & 73.42\% & mislabeling memorization\ 99.90\% \\
 & LDS                  & 99.91\% & 80.23\% & 3 random subsets acc.\ 69.32\%/69.80\%/68.60\% \\
\cmidrule(lr){1-5}
\multirow{6}{*}{\textbf{QNLI}}
 & Class Detection       & 84.29\% & 84.08\% & --- \\
 & TopKCardinality      & 84.29\% & 84.08\% & --- \\
 & ModelRandomization   & 84.29\% & 84.08\% & --- \\
 & Mislabeling Detection & 83.94\% & 84.79\% & mislabeling memorization\ 17.57\% \\
 & LDS                  & 84.29\% & 84.08\% & 3 random subsets acc.\ 83.31\%/83.51\%/83.47\% \\
 & Mixed Datasets        & 85.07\% & 84.47\% & adversarial dataset\ train/eval 99.35\%/98.91\% \\
\bottomrule
\end{tabularx}
\end{table}

\subsection{Metric Hyperparameters}\label{exp-bench-metrics}
This subsection reports the metric hyperparameters used in the benchmarks released with \appquanda. For all classification benchmarks, metric evaluation is performed on a randomly sampled subset of $1000$ examples from the corresponding evaluation dataset. Additional information about the benchmark datasets and the perturbations applied to them is reported in~\Cref{tab:bench-wrappers} separately.

\paragraph{LDS} The protocol uses $m=100$ random training subsets at subset ratio $\alpha=0.5$ and reports the Spearman rank correlation between predicted and observed effects. Results for different $m$ and $\alpha$ values are reported in~\Cref{bench_robustness}. The choices $m=100$ and $\alpha=0.5$ are common in recent LDS-based TDA evaluation \citep{park2023trak,bae2024training,sun2026enhancing}, enabling comparison of results with those papers. Larger $m$ gives a more precise correlation estimate, but training and evaluation costs scale linearly with $m$, as each subset needs a separate model. We sample train subsets of fixed size uniformly without replacement, as is common in the literature.

\paragraph{Class Detection, Subclass Detection, and Top-$k$ Cardinality} Class Detection and Subclass Detection report precision over the top $s=5$ attributed training points per query, i.e., the fraction of the top-$s$ proponents whose label matches the comparison target. Top-$k$ Cardinality reports the size of the union of the top-$k$ attributed sets across test queries, with $k=5$.

\paragraph{Shortcut Detection} Evaluation is restricted to non-shortcut-class test samples that the model classifies as class~$0$ despite their ground-truth label being different.

\paragraph{Mixed Datasets Separation} Since all samples from the adversarial dataset are assigned a single adversarial label ($0$), we also restrict the evaluation to test samples predicted as this class.

\subsection{Details About Fact-Tracing Benchmarks} \label{fact-tracing-details}
For the fact tracing benchmark, we follow the approach described in~\citep{chang2025scalable}. We use the T-REx dataset, which pairs short factual prompts (e.g.,
\textit{``The capital of France is:''}) with their answers (e.g.,
\textit{``Paris''}) and a set of evidence sentences that contain
the correct answer \citep{elsahar-etal-2018-rex}, built from the KILT-filtered T-REx dataset \citep{petroni-etal-2021-kilt}. T-REx pairs entity--relation triples with entailing Wikipedia sentences via distant supervision, validated by crowdsourcing. KILT drops facts with >20 evidence sentences and relations with <1000 facts, and merges facts sharing a subject--relation pair. \cite{chang2025scalable} further remove facts with multiple correct target IDs, write templates to turn the triples into prompts, and sample 5k facts balanced for fact frequency with common targets downsampled. 

We pretrain GPT-2 small (124M) from scratch on the union of the full T-REx
corpus and a 10\% sample of OpenWebText, holding out 5\% of OpenWebText
for validation. Training uses AdamW ($\beta_1{=}0.9$, $\beta_2{=}0.95$, weight decay $0.1$, gradient clipping at $1.0$) with a cosine learning-rate schedule that warms up over $500$ steps to a peak of $6\times10^{-4}$ and decays to $6\times10^{-5}$. We train for $30{,}000$ steps at context length $512$ with an effective batch of $512$ sequences. The pretrained model is 
fine-tuned on 600 samples of this set with the loss computed only on answer tokens, to ensure that the model outputs the answer in an expected format. Finally, we run inference on $4.3k$ samples from this set and
retain only those prompts the model answers correctly ($\approx 33\%$),
yielding the basis of the evaluation set used by the benchmark.

In the metric evaluation step, we build an entailment matrix matching each prompt to its entailing evidence, deduplicating matching passages. The three benchmarks (MRR, Recall@$k$, TailPatch) randomly sample $200$ prompts from the evaluation set with up to $5$ evidence passages per prompt at sequence length $128$, tokenized with the GPT-2 BPE tokenizer. Recall@$k$ uses $k=3$. TailPatch takes a training step on each of the top-$10$ retrieved passages with  AdamW at learning rate $6 \times 10^{-5}$, $\beta = (0.9, 0.95)$, and weight decay $0.1$, then measures the change in the answer sequence probability.

\subsection{Training Data Attribution Methods}\label{exp-explainers}

This subsection lists TDA methods used in each evaluation setting in~\Cref{benchmarking} and~\Cref{app:additional-results}, as well as their parameters. On AwA2 we include a similarity-based baseline (RepSim)~\citep{caruana_case-based_1999}, Representer Points (ReprPoints)~\citep{yeh2018representer}, TracInCP~\citep{pruthi2020estimating}, EK-FAC~\citep{grosse2023studyinglargelanguagemodel}, and TRAK with a single model checkpoint (TRAK-1)~\citep{park2023trak}. On QNLI, TracInCP becomes prohibitively expensive (due to the lack of caching in \texttt{Captum}'s TracInCPFast), so we restrict the comparison to TRAK, EK-FAC, RepSim, and ReprPoints. In the GPT-2 fact-tracing setting, attributions are computed only over the candidate set rather than the full training corpus; this rules out ReprPoints but allows us to additionally include the DataInf variant from \texttt{dattri}~\citep{kwon2024datainf}. To stay within our compute budget, we restrict TDA computation to a subset of layers per method. 

Throughout this subsection, RepSim, ArnoldiInf, and TracInCP are run via the \texttt{Captum} library~\citep{kokhlikyan_captum_2020}. TracInCP uses \texttt{Captum}'s TracInCPFast, restricted to the final linear layer and computed at five evenly spaced checkpoints over the course of training. ReprPoints uses the replication of the official repository accompanying~\citet{yeh2018representer}. TRAK-1 uses the official \texttt{TRAK} implementation~\citep{park2023trak} on the image and text classification settings and the TRAK attributor from \texttt{dattri}~\citep{dattri} on the language-modeling benchmarks. EK-FAC is run via \texttt{Kronfluence}~\citep{grosse2023studyinglargelanguagemodel} on the image and text classification, and via \texttt{dattri}~\citep{dattri} for language-modeling. DataInf~\citep{kwon2024datainf} is run via \texttt{dattri}. The remainder of this subsection reports the explainer configurations by setting. For each setting, we list the swept hyperparameters. Full per-configuration evaluation results and selected parameters are visualized in~\Cref{tda_hyperparam}.

\paragraph{MNIST}
RepSim uses cosine similarity, swept over the ReLU following the first linear layer, the second-to-last linear layer, and the final linear layer's input ReLU.
ArnoldiInf uses cross-entropy loss and Arnoldi dimension $200$, swept over $\{$final linear layer only, all three linear layers$\}$ $\times$ projection dimension $\in\{50,100\}$.
ReprPoints uses the final linear layer as classifier, swept over $\{$second-to-last linear layer, final linear layer's input ReLU$\}$ $\times$ feature $\ell_2$-normalization on/off.
TRAK-1 sweeps projection dimension in $\{512, 1024\}$.

\paragraph{CIFAR}
RepSim uses cosine similarity at the penultimate post-pooling features.
ArnoldiInf uses cross-entropy loss and Arnoldi dimension $200$, swept over $\{$final linear layer only, final convolutional block + final linear layer$\}$ $\times$ projection dimension $\in\{50, 100\}$.
ReprPoints uses the post-pooling features as input to a final linear classifier, swept over $\ell_2$-normalization on/off.
TRAK-1 sweeps projection dimension in $\{1024, 2048\}$.

\paragraph{AwA2}
RepSim uses cosine similarity at the post-average-pool penultimate features.
ReprPoints uses the post-pool penultimate features as input to the final linear classifier, swept over $\ell_2$-normalization on/off.
EK-FAC is computed over the last convolutional layer and the final linear layer.
TRAK-1 sweeps projection dimension in $\{1024, 2048, 4096\}$.

\paragraph{QNLI}
RepSim uses the BERT pooler-dropout output as the feature layer, swept over $\{$cosine, dot-product$\}$ similarity.
ReprPoints uses the same BERT pooler-dropout output as input to the final linear classifier head, swept over $\ell_2$-normalization on/off and randomization of initial method weights on/off.
TRAK-1 uses projection dimension $2048$ and no regularization.
EK-FAC is computed over the BERT pooler dense layer and the final linear classifier head.

\paragraph{T-REx}
RepSim uses the final-LayerNorm hidden states as the feature layer, swept over $\{$cosine, dot-product$\}$ similarity.
EK-FAC is computed over the attention and MLP projections of all twelve transformer blocks.
DataInf is computed over the two MLP projections (input and output) of the final transformer block, with regularization $\lambda=10^{-2}$ and $20\%$ of the training data used to estimate the empirical Fisher.
TRAK (via \texttt{dattri}) uses the same two projections, projection dimension $1024$, and regularization $\lambda=10^{-5}$.
\section{Additional Experimental Results} \label{app:additional-results}

This section presents additional empirical results: evaluation on the two smaller-scale ready-made settings (\Cref{app:extra_benchmark_results}), robustness analyses of the benchmark configurations (\Cref{bench_robustness}), and per-configuration sweep visualizations behind the TDA evaluation figures (\Cref{tda_hyperparam}).

\subsection{Benchmark Evaluation Results} \label{app:extra_benchmark_results}

\begin{figure*}[htb]
\centering
\includegraphics[width=\textwidth]{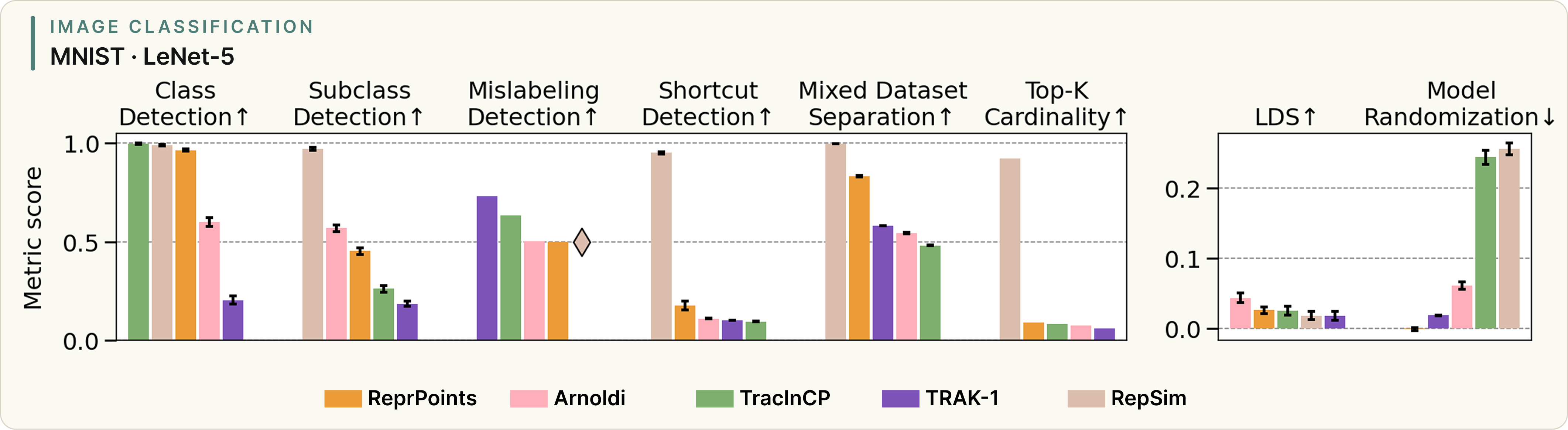}
\caption{Image-classification benchmarks (MNIST/LeNet-5). Conventions as in~\Cref{fig:eval-image-classification}. Random baselines (mean $\pm$ std over $41$ seeds): $0.100{\pm}0.004$ (Class), $0.099{\pm}0.004$ (Subclass), $0.500{\pm}0.001$ (Mislabeling), $0.094{\pm}0.000$ (Shortcut), $0.500{\pm}0.000$ (Mixed Datasets), $0.949{\pm}0.003$ (Top-$k$), $0.000{\pm}0.000$ (Model Randomization), and $0.000{\pm}0.003$ (LDS). These results mirror AwA2 evaluation scores: RepSim leads the similarity-rewarding benchmarks (Subclass, Shortcut, Mixed Datasets, Top-$k$), Mislabeling Detection is led by gradient-based methods, and LDS scores remain uniformly low.
}
\label{fig:eval-image-classification-mnist}
\end{figure*}

\begin{figure*}[t]
\centering
\includegraphics[width=\textwidth]{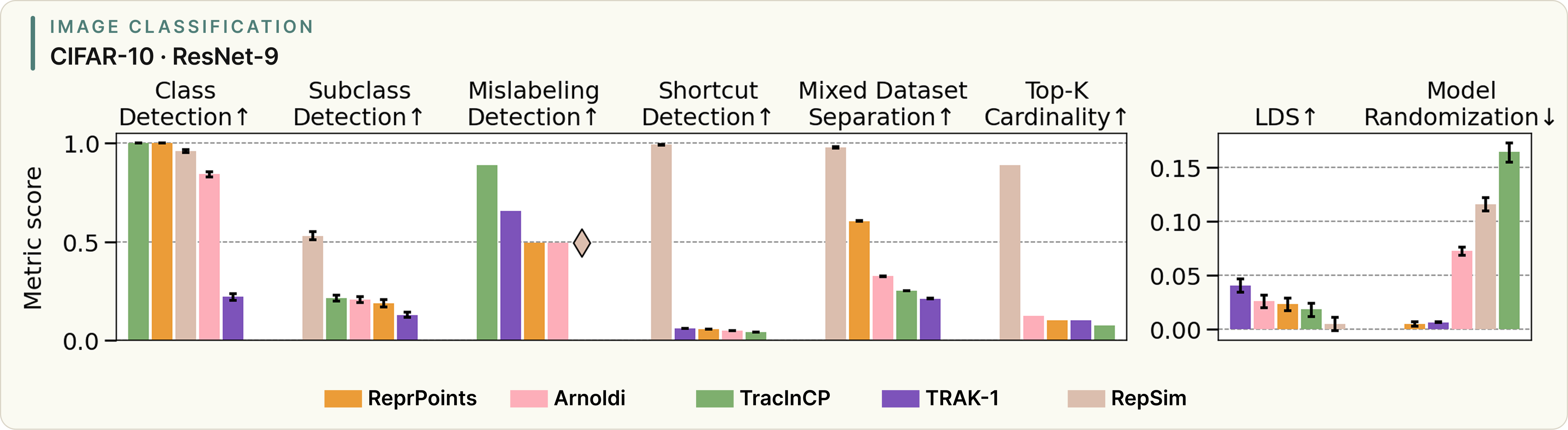}
\caption{Image-classification benchmarks (CIFAR-10/ResNet-9). Conventions as in~\Cref{fig:eval-image-classification}. Random baselines (mean $\pm$ std over $41$ seeds): $0.101{\pm}0.004$ (Class), $0.101{\pm}0.004$ (Subclass), $0.496{\pm}0.001$ (Mislabeling), $0.075{\pm}0.000$ (Shortcut), $0.227{\pm}0.000$ (Mixed Datasets), $0.939{\pm}0.003$ (Top-$k$), $0.000{\pm}0.000$ (Model Randomization), and $0.000{\pm}0.003$ (LDS). These results mirror AwA2 evaluation scores.
}
\label{fig:eval-image-classification-cifar}
\end{figure*}

This subsection extends the evaluation with results on the two additional ready-made \appquanda settings: MNIST with LeNet-5 in~\Cref{fig:eval-image-classification-mnist} and CIFAR-10 with ResNet-9 in~\Cref{fig:eval-image-classification-cifar}.

\subsection{Benchmark Robustness} \label{bench_robustness}

\begin{figure*}[ht!]
\centering
\includegraphics[width=0.8216981132 \textwidth]{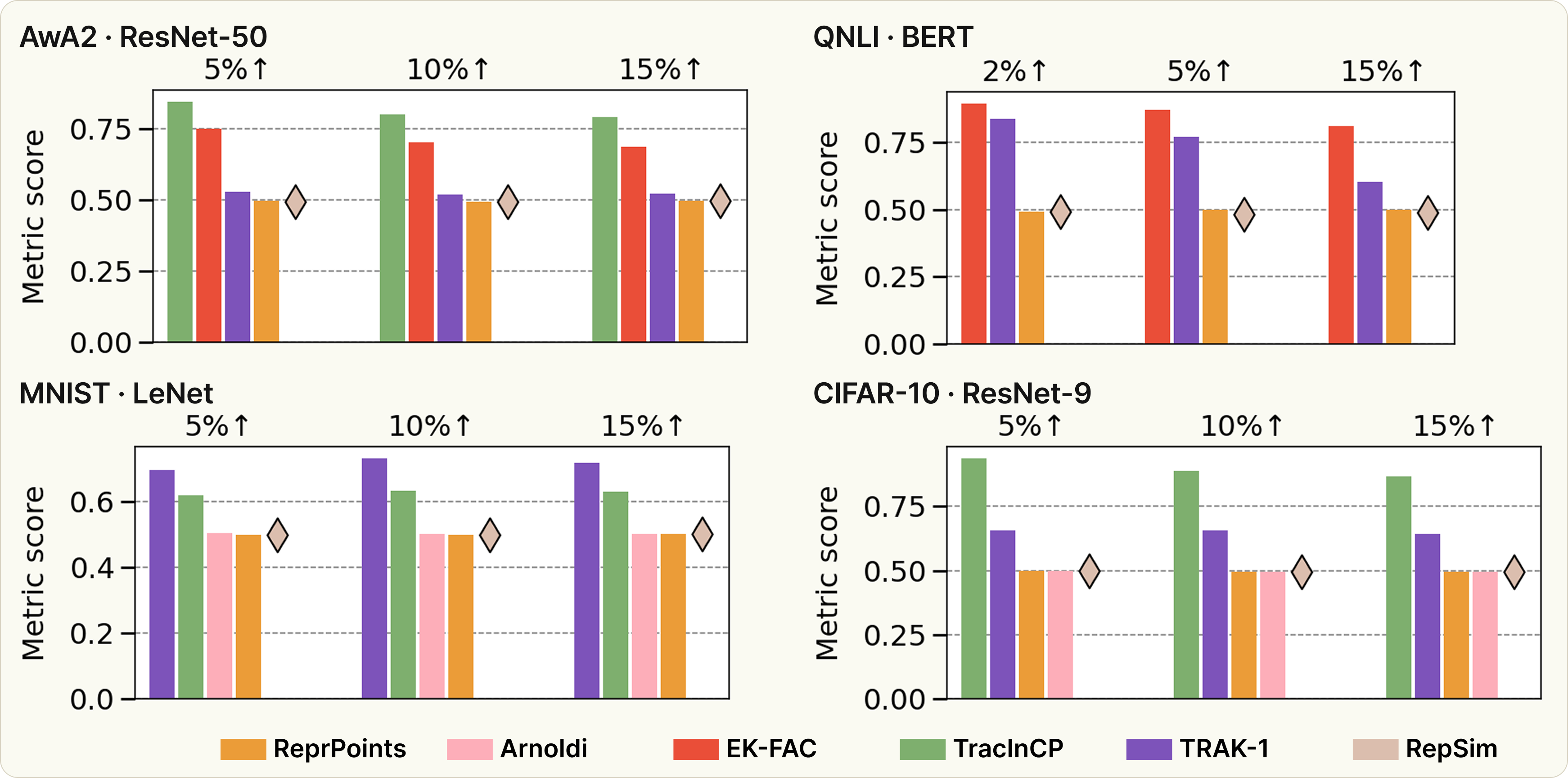}
\caption{Results across Mislabeling Detection benchmarks with varying percentages of corrupted data, extending the results reported in~\Cref{fig:eval-image-classification,fig:eval-text-classification,fig:eval-image-classification-mnist,fig:eval-image-classification-cifar}. Each setting is evaluated at 3 corruption rates; the rate used for the ready-made benchmarks is $10\%$ on \textbf{AwA2}, $2\%$ on \textbf{QNLI} (\Cref{tab:bench-wrappers}), and $10\%$ on \textbf{MNIST} and \textbf{CIFAR} (\Cref{tab:bench-wrappers}). Conventions as in~\Cref{fig:eval-image-classification}.}
\label{fig:mislabeling-percentages}
\end{figure*}

\begin{figure*}[ht!]
\centering

\includegraphics[width=0.6316037736 \textwidth]{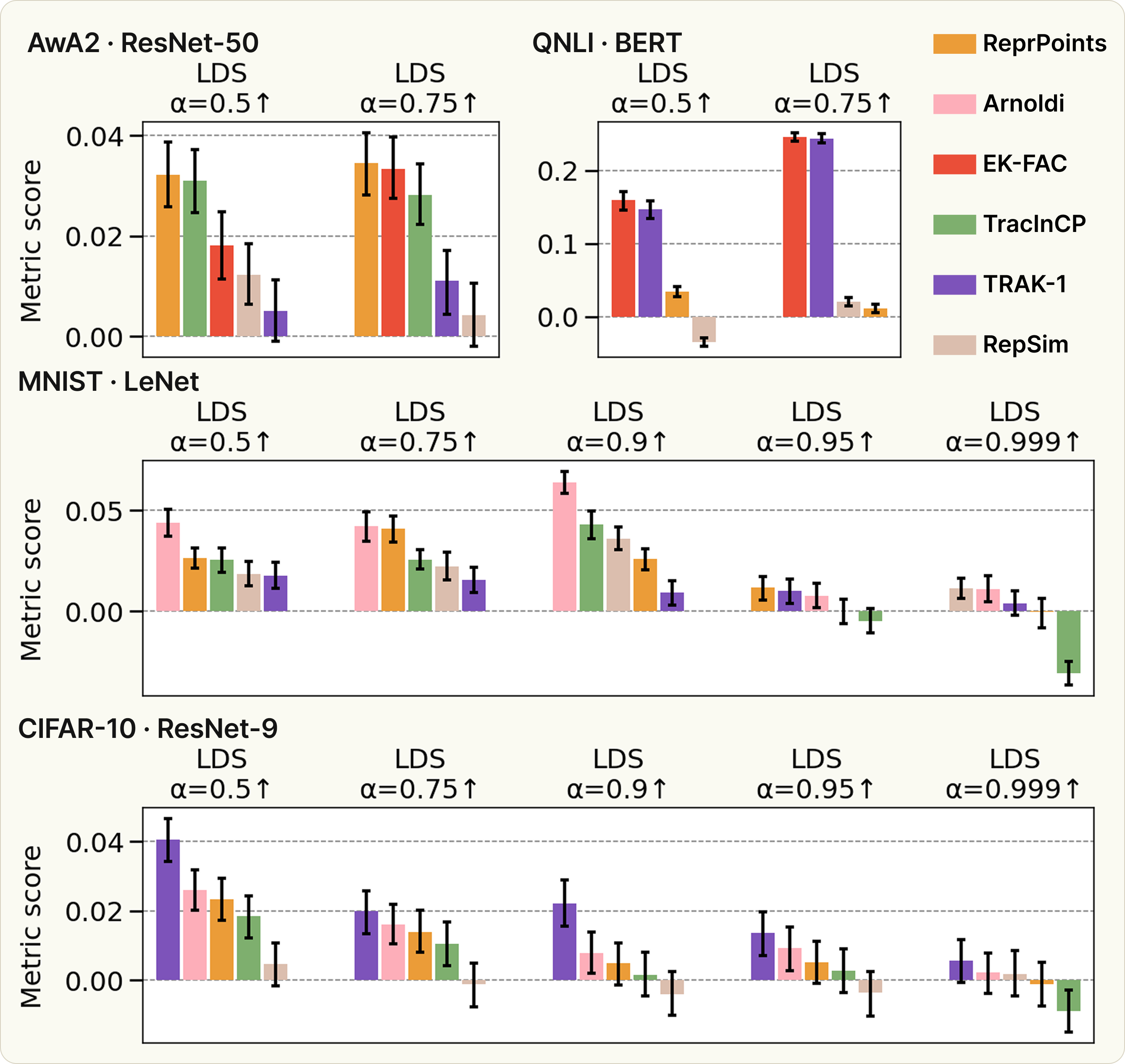}
\caption{LDS evaluation results across benchmarks with varying subset fraction $\alpha$, compared to $\alpha=0.5$ as reported in~\Cref{fig:eval-image-classification,fig:eval-text-classification,fig:eval-image-classification-mnist,fig:eval-image-classification-cifar}. Rankings are largely stable: on QNLI the ordering is unchanged for the well-performing TRAK-1 and EK-FAC, although these two methods improve markedly at $\alpha=0.75$; on AwA2, EK-FAC performs substantially better at $\alpha=0.75$. This is consistent with both methods being local counterfactual approximations around the full-data optimum, which are more accurate at larger subsets. On CIFAR, scores are near zero for $\alpha \geq 0.95$: as subsets near the full dataset, logit variance shrinks, and LDS becomes noise-dominated.}
\label{fig:lds-alpha}
\end{figure*}

This section examines the robustness of the benchmark results to the benchmarks' configuration parameters.~\Cref{fig:mislabeling-percentages} reports Mislabeling Detection results under varying percentages of mislabeled training samples. For Model Randomization, a single random model is used by default, as attribution cost scales linearly in the number of models. We have implemented an \texttt{n{\_}rand{\_}models} argument for Model Randomization for multi-model variance estimation. We report mean $\pm$ std over 10 random models in~\Cref{tab:mrand-n}. For LDS, we vary the training-subset fraction $\alpha$ in~\Cref{fig:lds-alpha} and the number of retrained subset models $m$ used to compute the score in~\Cref{tab:lds-m}.

\begin{table}[t]
\centering
\caption{Model Randomization computed with a single randomized model, as used in the benchmark evaluations reported in~\Cref{fig:eval-image-classification,fig:eval-image-classification-mnist,fig:eval-image-classification-cifar}, compared against 10 independently randomized models (mean $\pm$ std over the scores per random model). Single-model scores generally align with the 10-model means, and the single-model ranking matches the 10-model ranking in every setting:}
\label{tab:mrand-n}
\begin{tabular}{@{}llcc@{}}
\toprule
Setting & Explainer & 1 model & 10 models \\
\midrule
\multirow{5}{*}{\textbf{MNIST}}
&RepSim & $0.255$ & $0.267 \pm 0.044$ \\
&ReprPoints & $-0.001$ & $-0.001 \pm 0.004$ \\
&TracInCP & $0.244$ & $0.239 \pm 0.023$ \\
&ArnoldiInf & $0.061$ & $0.058 \pm 0.004$ \\
&TRAK-1 & $0.019$ & $0.018 \pm 0.001$ \\
\midrule
\multirow{5}{*}{\textbf{CIFAR}}
&RepSim & $0.116$ & $0.152 \pm 0.017$ \\
&ReprPoints & $0.005$ & $0.002 \pm 0.005$ \\
&TracInCP & $0.165$ & $0.164 \pm 0.002$ \\
&ArnoldiInf & $0.073$ & $0.069 \pm 0.006$ \\
&TRAK-1 & $0.006$ & $0.006 \pm 0.000$ \\
\midrule
\multirow{5}{*}{\textbf{AwA2}}
&RepSim & $0.171$ & $0.181 \pm 0.009$ \\
&ReprPoints & $-0.004$ & $0.000 \pm 0.004$ \\
&TracInCP & $0.085$ & $0.058 \pm 0.017$ \\
&TRAK-1 & $0.004$ & $0.003 \pm 0.000$ \\
&EK-FAC & $0.018$ & $0.019 \pm 0.001$ \\
\bottomrule
\end{tabular}
\end{table}

\begin{table}[ht!]
\centering
\caption{LDS computed from a draw of $m$ subset models out of 100, as used in the benchmark evaluations reported in~\Cref{fig:eval-image-classification,fig:eval-text-classification,fig:eval-image-classification-mnist,fig:eval-image-classification-cifar}. For $m<100$, cells are mean $\pm$ std over 20 random draws of $m$; $m=100$ is the full benchmark. Draws are shared across explainers. In settings where the benchmark scores are non-negligible, the results are noisy at $m=10$ and saturate by $m=50$, indicating that $m=100$ suffices here.}
\label{tab:lds-m}
\begin{tabular}{@{}llcccc@{}}
\toprule
Setting & Explainer & $m=10$ & $m=25$ & $m=50$ & $m=100$ \\
\midrule
\multirow{5}{*}{\textbf{MNIST}}
 & RepSim & $0.014 \pm 0.063$ & $0.021 \pm 0.036$ & $0.018 \pm 0.024$ & $0.018$ \\
 & ReprPoints & $0.031 \pm 0.070$ & $0.032 \pm 0.037$ & $0.023 \pm 0.021$ & $0.026$ \\
 & TracInCP & $0.002 \pm 0.063$ & $0.021 \pm 0.037$ & $0.026 \pm 0.020$ & $0.025$ \\
 & ArnoldiInf & $0.015 \pm 0.060$ & $0.038 \pm 0.036$ & $0.039 \pm 0.024$ & $0.044$ \\
 & TRAK-1 & $0.015 \pm 0.014$ & $0.020 \pm 0.006$ & $0.018 \pm 0.004$ & $0.018$ \\
\midrule
\multirow{5}{*}{\textbf{CIFAR}}
 & RepSim & $0.006 \pm 0.013$ & $0.007 \pm 0.007$ & $0.006 \pm 0.005$ & $0.005$ \\
 & ReprPoints & $0.019 \pm 0.010$ & $0.026 \pm 0.007$ & $0.023 \pm 0.003$ & $0.023$ \\
 & TracInCP & $0.019 \pm 0.011$ & $0.019 \pm 0.006$ & $0.018 \pm 0.004$ & $0.018$ \\
 & ArnoldiInf & $0.024 \pm 0.012$ & $0.027 \pm 0.007$ & $0.026 \pm 0.004$ & $0.026$ \\
 & TRAK-1 & $0.034 \pm 0.011$ & $0.041 \pm 0.005$ & $0.040 \pm 0.003$ & $0.040$ \\
\midrule
\multirow{5}{*}{\textbf{AwA2}}
 & RepSim & $0.010 \pm 0.016$ & $0.013 \pm 0.007$ & $0.011 \pm 0.005$ & $0.012$ \\
 & ReprPoints & $0.031 \pm 0.014$ & $0.030 \pm 0.008$ & $0.031 \pm 0.005$ & $0.032$ \\
 & TracInCP & $0.031 \pm 0.012$ & $0.029 \pm 0.007$ & $0.033 \pm 0.003$ & $0.031$ \\
 & TRAK-1 & $0.006 \pm 0.010$ & $0.006 \pm 0.006$ & $0.006 \pm 0.003$ & $0.005$ \\
 & EK-FAC & $0.015 \pm 0.010$ & $0.019 \pm 0.006$ & $0.018 \pm 0.003$ & $0.018$ \\
\midrule
\multirow{5}{*}{\textbf{QNLI}}
 & RepSim & $-0.026 \pm 0.052$ & $-0.038 \pm 0.031$ & $-0.028 \pm 0.025$ & $-0.035$ \\
 & ReprPoints & $0.044 \pm 0.045$ & $0.025 \pm 0.029$ & $0.030 \pm 0.016$ & $0.034$ \\
 & TRAK-1 & $0.137 \pm 0.007$ & $0.141 \pm 0.006$ & $0.145 \pm 0.004$ & $0.146$ \\
 & EK-FAC & $0.148 \pm 0.010$ & $0.155 \pm 0.005$ & $0.158 \pm 0.004$ & $0.159$ \\
\bottomrule
\end{tabular}
\end{table}

\clearpage
\subsection{Explainer Hyperparameter Selection} \label{tda_hyperparam}

\begin{figure*}[ht]
\centering
\includegraphics[width=0.6816037736 \textwidth]{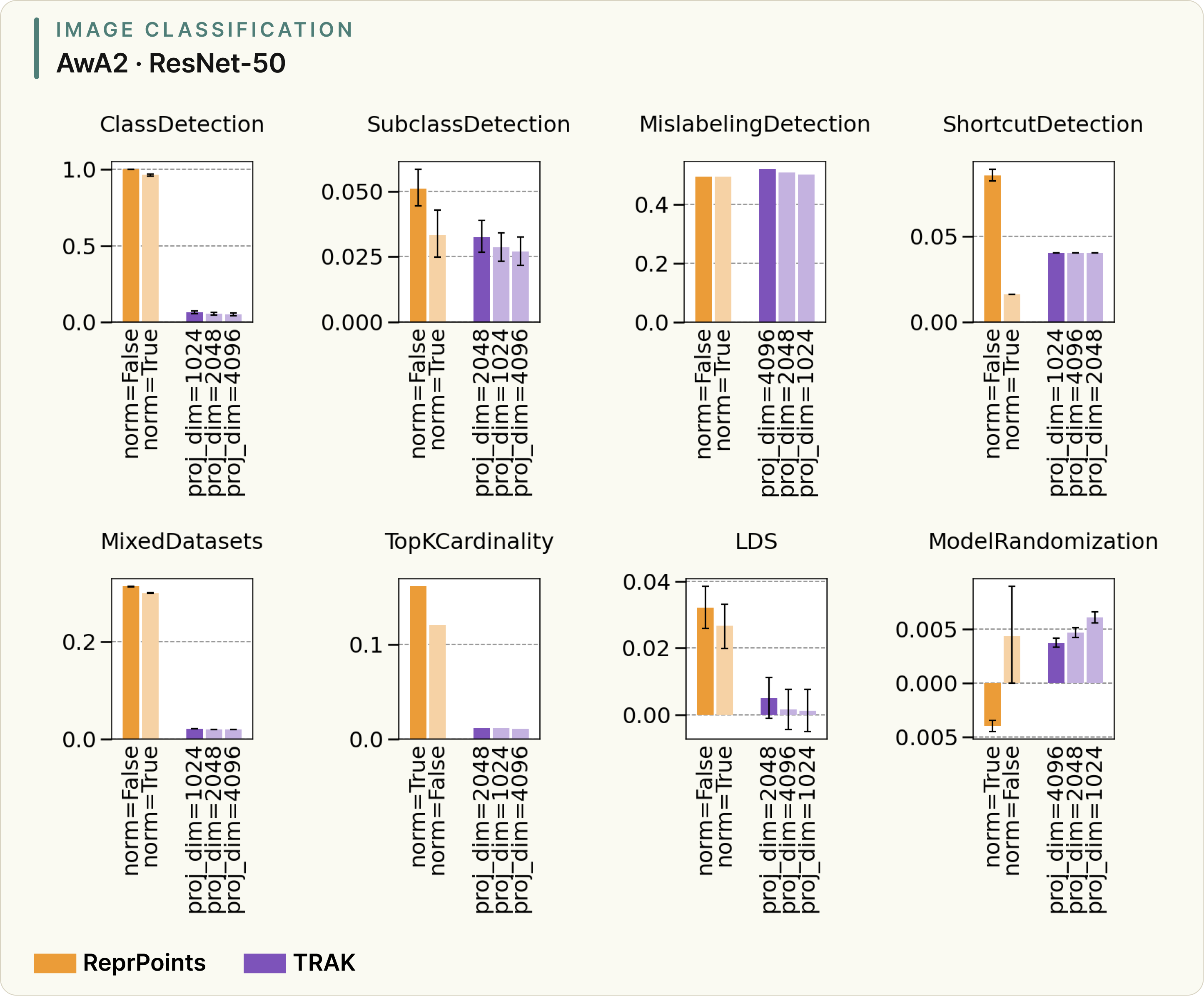}
\caption{Hyperparameter sweep on AwA2/ResNet-50. For each benchmark and
attribution method, every explored hyperparameter configuration is shown
as a separate bar; the configuration selected for the main results is
drawn in the solid method colour, and the remaining sweep runs in a
lighter shade. Error bars give the 95\% confidence interval where
available. The same conventions apply
to~\Cref{fig:hyperparam-qnli,fig:hyperparam-mnist,fig:hyperparam-cifar}.}
\label{fig:hyperparam-awa}
\end{figure*}

\begin{figure*}[ht!]
\centering
\includegraphics[width=0.8270141509\textwidth]{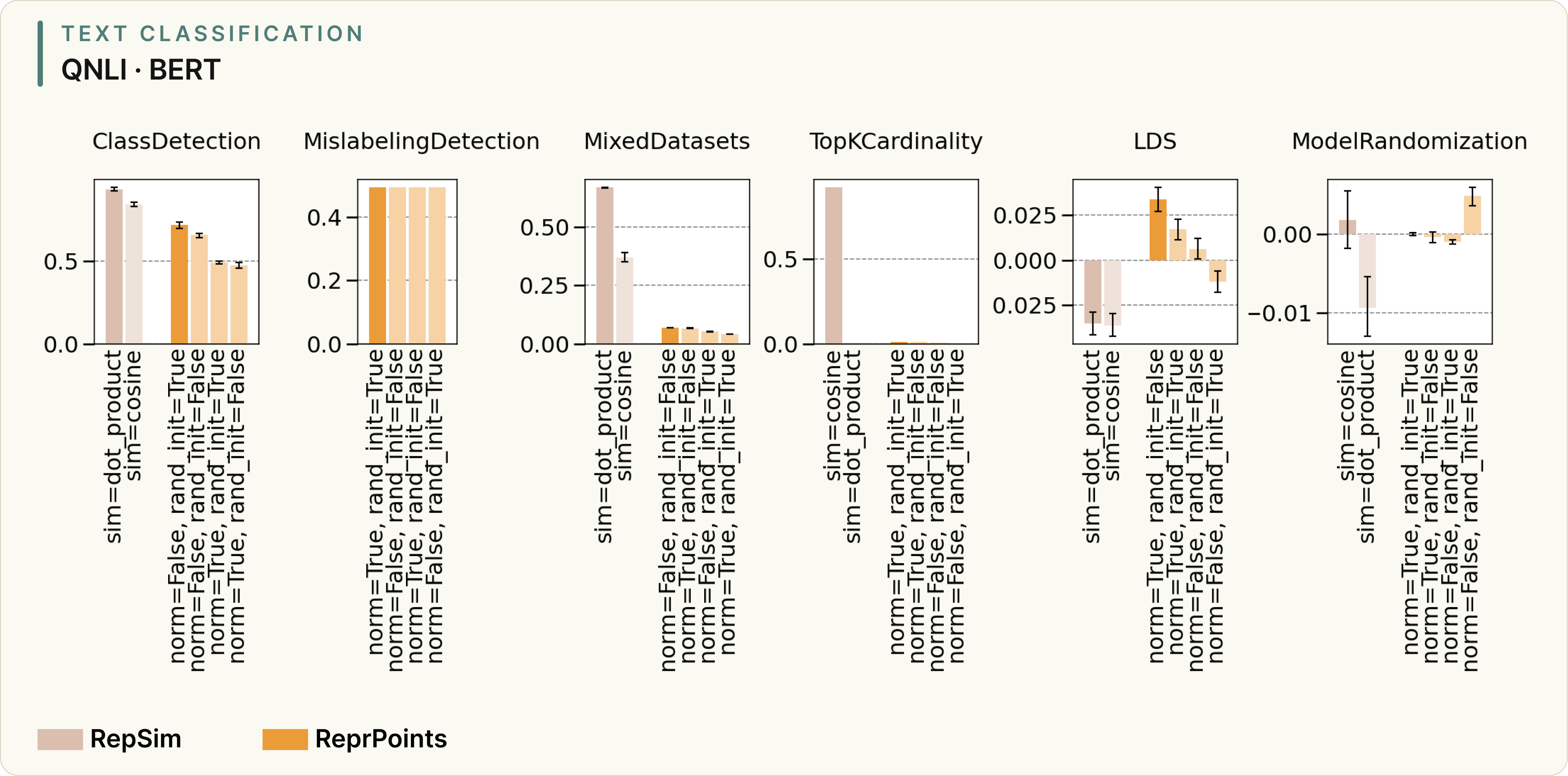}
\caption{Hyperparameter sweep on QNLI/BERT. Conventions as in
~\Cref{fig:hyperparam-awa}.}
\label{fig:hyperparam-qnli}
\end{figure*}

\begin{figure*}[t]
\centering
\includegraphics[width=0.8350943396\textwidth]{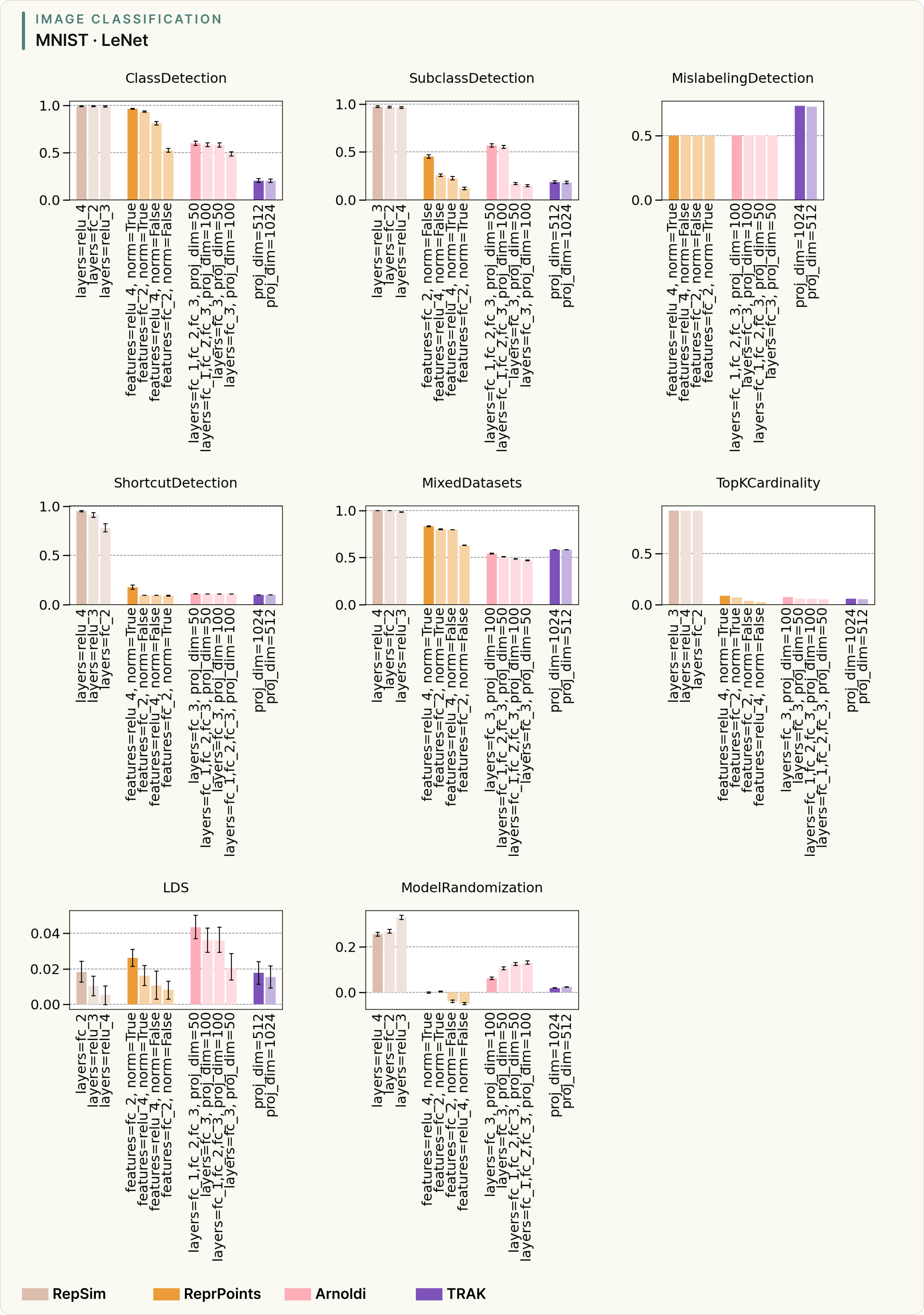}
\caption{Hyperparameter sweep on MNIST/LeNet-5. Conventions as in
~\Cref{fig:hyperparam-awa}.}
\label{fig:hyperparam-mnist}
\end{figure*}

\begin{figure*}[t]
\centering
\includegraphics[width=0.8160377358 \textwidth]{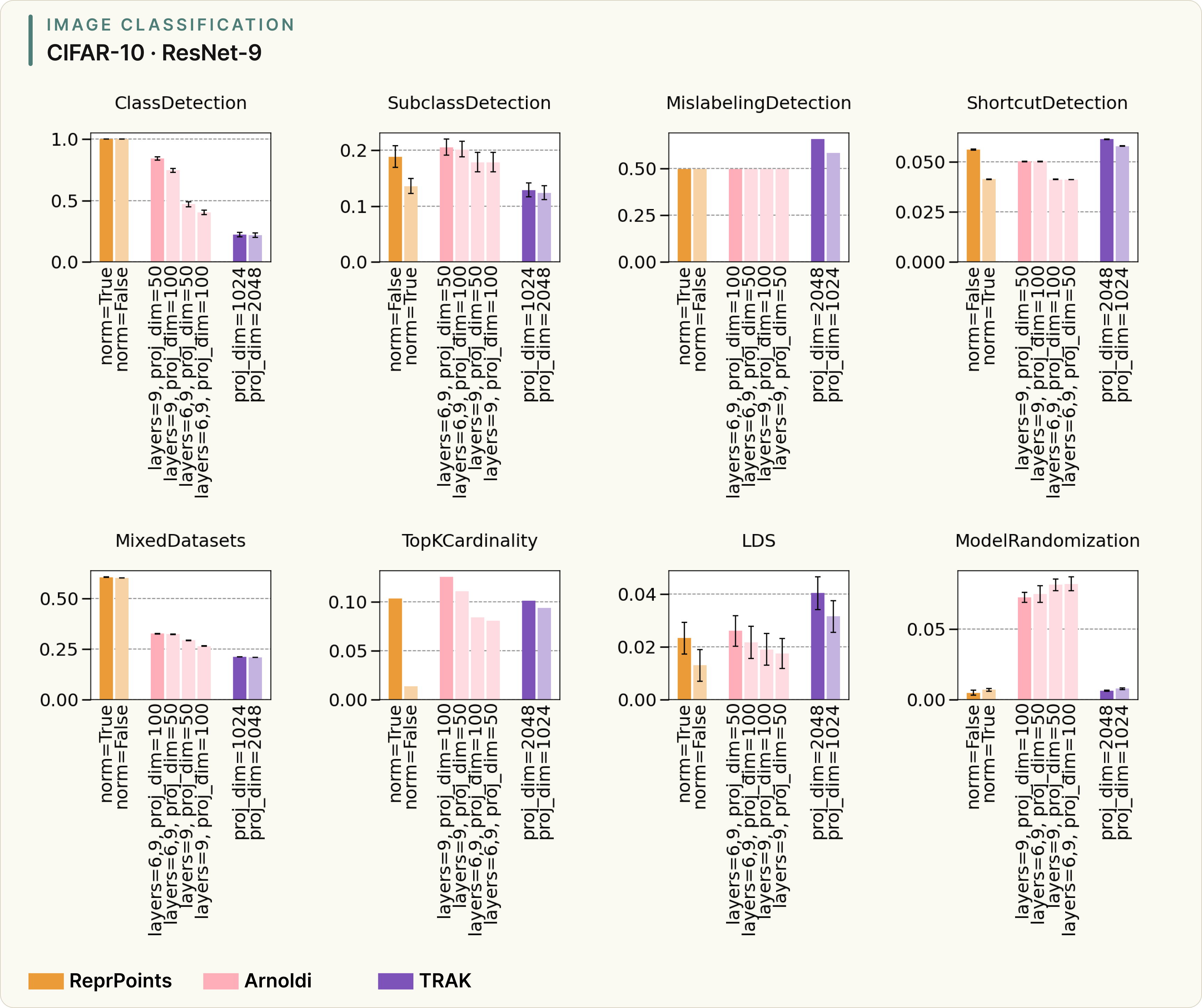}
\caption{Hyperparameter sweep on CIFAR-10/ResNet-9. Conventions as in
~\Cref{fig:hyperparam-mnist}.}
\label{fig:hyperparam-cifar}
\end{figure*}

We present the full hyperparameter sweeps behind the TDA evaluation results, allowing the reader to assess each method's sensitivity to its hyperparameters and the robustness of the relative method ordering across the sweep grid. Specifically, the bar charts in~\Cref{fig:eval-image-classification,fig:eval-text-classification,fig:eval-image-classification-mnist,fig:eval-image-classification-cifar} report, for each method and each metric, the score of the best-performing configuration. \Cref{fig:hyperparam-mnist,fig:hyperparam-cifar,fig:hyperparam-awa,fig:hyperparam-qnli} show every evaluated configuration as a separate bar, with the configuration selected for those bar charts drawn in the method's solid color and all others in a lighter shade. The sweep grids are listed in~\Cref{exp-explainers}. We omit the fact-tracing setting from this analysis, as no hyperparameter sweep was performed in that configuration.

\clearpage

\section{Compute Resources} \label{compute_resources}

All experiments were conducted using a workstation with 2 \texttimes{} 24 GB NVIDIA\textregistered{} RTX 4090 GPUs and a compute cluster with 4 \texttimes{} 40 GB NVIDIA\textregistered{} A100 GPUs, with each individual run performed on a single GPU. As a rough order-of-magnitude estimate, source-model training takes minutes on \textbf{MNIST}/\textbf{CIFAR}, a few hours on \textbf{AwA2} and \textbf{QNLI}, and roughly a day for the \textbf{T-REx} pretraining run. Per-attribution evaluation ranges from minutes in smaller settings (RepSim, ReprPoints) to a few hours (TRAK, EK-FAC, TracInCP) on the larger settings. Including the full hyperparameter sweeps in~\Cref{exp-explainers} and the LDS subset retrainings, we estimate the total compute spent on the reported results at roughly $500$ GPU-hours, including the budget consumed by preliminary and discarded runs.

\clearpage
\section*{NeurIPS Paper Checklist}

\begin{enumerate}

\item {\bf Claims}
    \item[] Question: Do the main claims made in the abstract and introduction accurately reflect the paper's contributions and scope?
    \item[] Answer: \answerYes{}
    \item[] Justification: In the abstract, we claim to have developed a library that facilitates TDA attribution. We provide the link to the repository and the documentation, and describe our software contribution throughout \Cref{library,metrics_desc}. The claims about TDA methods trade-offs are substantiated in \Cref{benchmarking}.

\item {\bf Limitations}
    \item[] Question: Does the paper discuss the limitations of the work performed by the authors?
    \item[] Answer: \answerYes{}
    \item[] Justification: We discuss the limitations in \Cref{conclusion}.

\item {\bf Theory assumptions and proofs}
    \item[] Question: For each theoretical result, does the paper provide the full set of assumptions and a complete (and correct) proof?
    \item[] Answer: \answerNA{}
    \item[] Justification: We provide no theoretical results in this paper.

    \item {\bf Experimental result reproducibility}
    \item[] Question: Does the paper fully disclose all the information needed to reproduce the main experimental results of the paper to the extent that it affects the main claims and/or conclusions of the paper (regardless of whether the code and data are provided or not)?
    \item[] Answer: \answerYes{}
    \item[] Justification: The experimental details necessary to reproduce our results, including datasets, models, methods, metrics, and parameters we use, are provided throughout the main text and in the Appendix (\Cref{exp_details}).

\item {\bf Open access to data and code}
    \item[] Question: Does the paper provide open access to the data and code, with sufficient instructions to faithfully reproduce the main experimental results, as described in supplemental material?
    \item[] Answer: \answerYes{}
    \item[] Justification: We provide the link to the repository. The \texttt{scripts} folder in the repository contains all the scripts necessary to reproduce the results. All data and model assets are publicly available through HuggingFace.

\item {\bf Experimental setting/details}
    \item[] Question: Does the paper specify all the training and test details (e.g., data splits, hyperparameters, how they were chosen, type of optimizer) necessary to understand the results?
    \item[] Answer: \answerYes{}
    \item[] Justification: We specify and justify the training and test details, including dataset splits, hyperparameters, and optimizer types in \Cref{exp_details}.

\item {\bf Experiment statistical significance}
    \item[] Question: Does the paper report error bars suitably and correctly defined or other appropriate information about the statistical significance of the experiments?
    \item[] Answer: \answerYes{}
    \item[] Justification: For each benchmark, we report $95\%$ confidence intervals around the metric score, obtained by resampling the per-test-sample scores returned by the metric with $1000$ ($200$ for fact-tracing setting) bootstrap replicates, shown as error bars where per-sample resampling is meaningful. For the Random explainer baseline, we report mean $\pm$ standard deviation of the metric score across $41$ independent random seeds ($\texttt{seed}\in\{0,\dots,40\}$). These are reported in figures of \Cref{benchmarking} and \Cref{app:additional-results}.

\item {\bf Experiments compute resources}
    \item[] Question: For each experiment, does the paper provide sufficient information on the computer resources (type of compute workers, memory, time of execution) needed to reproduce the experiments?
    \item[] Answer: \answerYes{}
    \item[] Justification: The computational resources are detailed in \Cref{compute_resources}.
    
\item {\bf Code of ethics}
    \item[] Question: Does the research conducted in the paper conform, in every respect, with the NeurIPS Code of Ethics \url{https://neurips.cc/public/EthicsGuidelines}?
    \item[] Answer: \answerYes{}
    \item[] Justification: This paper complies with the NeurIPS Code of Ethics.

\item {\bf Broader impacts}
    \item[] Question: Does the paper discuss both potential positive societal impacts and negative societal impacts of the work performed?
    \item[] Answer: \answerYes{}
    \item[] Justification: The potential positive impact is described in \Cref{conclusion}; we do not expect our library to have a negative societal impact.
    
\item {\bf Safeguards}
    \item[] Question: Does the paper describe safeguards that have been put in place for responsible release of data or models that have a high risk for misuse (e.g., pre-trained language models, image generators, or scraped datasets)?
    \item[] Answer: \answerNA{}
    \item[] Justification: We do not release assets at a high risk of misuse, as most models have been trained on toy settings, and are all under 125M parameters.

\item {\bf Licenses for existing assets}
    \item[] Question: Are the creators or original owners of assets (e.g., code, data, models), used in the paper, properly credited and are the license and terms of use explicitly mentioned and properly respected?
    \item[] Answer: \answerYes{}
    \item[] Justification: We have specified all the assets used in our benchmarks, the links to the sources as well as the license details in \Cref{exp-datasets}.

\item {\bf New assets}
    \item[] Question: Are new assets introduced in the paper well documented and is the documentation provided alongside the assets?
    \item[] Answer: \answerYes{}
    \item[] Justification: We introduce benchmark assets, which include trained benchmark models and metadata. These are documented in this paper (\Cref{exp_details}) as well as in the repository documentation.

\item {\bf Crowdsourcing and research with human subjects}
    \item[] Question: For crowdsourcing experiments and research with human subjects, does the paper include the full text of instructions given to participants and screenshots, if applicable, as well as details about compensation (if any)? 
    \item[] Answer: \answerNA{}
    \item[] Justification: This paper does not involve crowdsourcing and research with human subjects.

\item {\bf Institutional review board (IRB) approvals or equivalent for research with human subjects}
    \item[] Question: Does the paper describe potential risks incurred by study participants, whether such risks were disclosed to the subjects, and whether Institutional Review Board (IRB) approvals (or an equivalent approval/review based on the requirements of your country or institution) were obtained?
    \item[] Answer: \answerNA{}
    \item[] Justification: This paper does not involve crowdsourcing and research with human subjects.

\item {\bf Declaration of LLM usage}
    \item[] Question: Does the paper describe the usage of LLMs if it is an important, original, or non-standard component of the core methods in this research? Note that if the LLM is used only for writing, editing, or formatting purposes and does \emph{not} impact the core methodology, scientific rigor, or originality of the research, declaration is not required.
    \item[] Answer: \answerNA{}
    \item[] Justification: LLMs were not used as part of the core methodology or for any important, original, or non-standard components of this work.

\end{enumerate}

\end{document}